\documentclass[lettersize,journal]{IEEEtran}
\usepackage{amsmath,amsfonts}
\usepackage{algorithmic}
\usepackage{array}
\usepackage[caption=false,font=normalsize,labelfont=sf,textfont=sf]{subfig}
\usepackage{textcomp}
\usepackage{stfloats}
\usepackage{url}
\usepackage{verbatim}
\usepackage{graphicx}

\usepackage{amsfonts}
\usepackage{amssymb}
\usepackage{bm}
\usepackage{multirow}
\usepackage{diagbox}
\usepackage[ruled]{algorithm2e}
\usepackage{color}
\newtheorem{remark}{Remark}

\usepackage{cite}
\makeatletter
\let\NAT@parse\undefined
\makeatother
\usepackage[]{hyperref}

\def\BibTeX{{\rm B\kern-.05em{\sc i\kern-.025em b}\kern-.08em
    T\kern-.1667em\lower.7ex\hbox{E}\kern-.125emX}}
\usepackage{balance}
\begin{document}
\title{Optimizing Control-Friendly Trajectories with Self-Supervised Residual Learning}
\author{Kexin Guo$^{*}$, Zihan Yang$^{*}$, Yuhang Liu, Jindou Jia$^{\dagger}$ and Xiang Yu$^{\dagger}$
\thanks{$^{*}$Equal contribution.}
\thanks{$^{\dagger}$Corresponding authors.}
\thanks{Kexin Guo is with the School of Aeronautic Science and Engineering, Beihang University, 
        37 Xueyuan Road, Beijing, 100191, China, and also with the Hangzhou Innovation Institute, 
        Beihang University, Changhe Street, Hangzhou, 310051, China.
        }
\thanks{Zihan Yang and Yuhang Liu are with the School of Aeronautic Science and Engineering, 
        Beihang University, 37 Xueyuan Road, Beijing, 100191, China.
        }
\thanks{Jindou Jia is with the School of Automation Science and Electrical Engineering, Beihang University, 
        37 Xueyuan Road, Beijing, 100191, China.
}
\thanks{Xiang Yu is with the School of Automation Science and Electrical Engineering, Beihang University, 
        37 Xueyuan Road, Beijing, 100191, China, and also with the Hangzhou Innovation Institute, 
        Beihang University, Changhe Street, Hangzhou, 310051, China.
}
}

\markboth{IEEE TRANSACTIONS ON XXX,~Vol.~XX, No.~XX, XXXX}%
{Optimizing Control-Friendly Trajectories\\ with Self-Supervised Residual Learning}

\maketitle

\begin{abstract}
  Real-world physics can only be analytically modeled with a certain level of
  precision for modern intricate robotic systems.
  As a result, tracking aggressive trajectories accurately could be challenging
  due to the existence of residual physics during controller synthesis.
  This paper presents a self-supervised residual learning and trajectory optimization
  framework to address the aforementioned challenges.
  At first, unknown dynamic effects on the closed-loop model are learned 
  and treated as residuals of the nominal dynamics, jointly forming a hybrid model.
  We show that learning with analytic gradients 
  can be achieved using only trajectory-level data while enjoying 
  accurate long-horizon prediction with an arbitrary integration step size.
  Subsequently, a trajectory optimizer is developed to compute
  the optimal reference trajectory with the residual physics along it minimized.
  It ends up with trajectories that are friendly to the following control level.
  The agile flight of quadrotors illustrates that by utilizing the hybrid dynamics, 
  the proposed optimizer outputs aggressive motions that can be precisely tracked.
  \end{abstract}

\begin{IEEEkeywords}
Residual Physics, Self-Supervised Learning, Trajectory Optimization for Control
\end{IEEEkeywords}


\section{Introduction}
\IEEEPARstart{G}{eneral} model-based approaches for robotic systems usually
decompose the task execution into a
planning stage and a control stage \cite{songReachingLimitAutonomous2023}.
Such setup has been widely applied to the quadrotor systems
with noticeable works
\cite{sunComparativeStudyNonlinear2022,
jiaEVOLVEROnlineLearning2024}.
However, in applications where aggressive planning objectives are required, 
there are few guarantees of accurate tracking control due to residual physics
\cite{sunComparativeStudyNonlinear2022}.

The challenges of accurate tracking are typically 
delegated to the control level.
Nevertheless, in aggressive motions, 
the unmodelled physics and model mismatches
heavily contribute to 
the degradation of tracking performance.
Currently, learning-based approaches 
have been developed to handle these effects, which have achieved remarkable results 
\cite{tase_learnCtrl,torrenteDataDrivenMPCQuadrotors2021,salzmannRealtimeNeuralMPCDeep2023}.
However, a promising result usually comes at the cost of
a complicated controller structure with onerous tuning 
\cite{richardsAdaptiveControlOrientedMetaLearningNonlinear2021, jiaEVOLVEROnlineLearning2024} or 
heavier onboard computation \cite{salzmannRealtimeNeuralMPCDeep2023}.
Fine-tuning tasks
\cite{romeroModelPredictiveContouring2022, loquercioAutoTuneControllerTuning2022, taoDiffTuneMPCClosedLoopLearning2023}
of controllers could be complicated and time-consuming to achieve promising tracking performance.
Therefore, both the synthesis and fine-tuning tasks of controllers remain non-trivial with difficulties.

Aware of the problems mentioned above, different from previous schemes that handle uncertainty in the control level,
we enhance the control performance by trajectory optimization.
The basic idea is to optimize a trajectory that is friendly to the following control level.
We name it \textit{control-friendly} trajectory optimization,
aiming at minimizing the residual physics that 
is neglected or unable to be treated during controller synthesis,
and thus enhancing the tracking performance of the controller.
Firstly, a regressor is augmented with a nominal closed-loop model
for capturing the residual physics, forming a hybrid model.
Such settings bring us a self-supervised learning framework
for capturing the residual physics free from labels 
of control inputs and state derivatives
that are usually noisy or unavailable in real-world applications,
which supervised learning methods \cite{jiaEVOLVEROnlineLearning2024,
torrenteDataDrivenMPCQuadrotors2021, bauersfeldNeuroBEMHybridAerodynamic2021} are delicate with.
The residual effect that causes the difference between
model rollouts and real-world trajectory data can be then learned 
by computing the gradient of the regressor parameters concerning the difference.
Besides, a stable integration of the predicting dynamics can be achieved
for stable prediction with an arbitrary step size, which is favored
by the long-horizon usage of the model, such as trajectory optimization.
We also show that the analytic gradients could be derived via
optimal control theory.
Next, a trajectory optimization 
is defined for residual minimization, which directly outputs the optimal reference commands
that favor the trajectory tracking with only baseline controllers.
By applying sparse formulation and warm-starting techniques
for the numerical optimization problem, the proposed trajectory optimizer
scales well to a high-dimensional quadrotor model.

Instead of reconfiguring the controllers with learned models, our approach
remains the original controller structure and utilizes the learned model
to favor the minimum-residual motion planning. 
It is suitable for various kinds of robotic applications with less or unauthorized labor on the deployment of the controllers.
To recap, our main contributions are listed as follows:

\begin{itemize}
\item[$\bullet$] A self-supervised learning framework for
capturing residual physics, which is free
from noisy labels with stable long-horizon prediction.
\item[$\bullet$] A minimum-residual trajectory optimization for
generating control-friendly reference commands, reducing the effects
of residual physics that is unknown during controller synthesis.
\end{itemize}

The outline of this letter is as follows.
Section II reviews the related works.
Section III provides the preliminaries.
Section IV goes into the details on the self-supervised learning and minimum-residual
trajectory optimization.
Section V presents simulations and numerical studies.
Section VI shows the results from real-world experiments.
Section VII concludes this article with a discussion. 

\section{Related Work}
\subsection{Learning-based Dynamics}
The concept of dynamics learning originally relates 
to system identification approaches.
Various methods have been developed from the learning aspect.
Koopman theory-based approaches lift the nonlinear inputs and
outputs (IOs) into a higher dimensional space in which linear mappings exist
\cite{brunton2021modernkoopmantheorydynamical, jackson2023dataefficientmodellearningcontrol, jiaEVOLVEROnlineLearning2024}.
However, prior knowledge of lifting function selection is required.
Deep neural networks 
are effective tools for dynamics learning with fewer priors
\cite{tase_nn_NARX, tase_nn_MPC,
salzmannRealtimeNeuralMPCDeep2023,
savioloPhysicsInspiredTemporalLearning2022a,nnMPC_AutonomousDriving2022,MillimeterLevel_PegInHole_2024},
but direct learning could be data-hungry
and only the loss of one-step prediction is considered.
Gaussian processes
can provide direct insights into model uncertainty
\cite{torrenteDataDrivenMPCQuadrotors2021, sun2024SafeStabilizationModel},
while the computational complexity increases with the amount of training data.
The idea of residual learning 
has been widely introduced to 
achieve higher training performance with hybrid models.
Hybrid models have been established in the noticeable works for quadrotors
with aerodynamic effects treated as residual physics 
\cite{salzmannRealtimeNeuralMPCDeep2023, bauersfeldNeuroBEMHybridAerodynamic2021},
where the existence of a nominal model or extra first-principle components
aids to enhance the generalizability and data efficiency.
A structural decomposition of the quadrotor dynamics is introduced in \cite{bansalLearningQuadrotorDynamics2016}
where the residual physics is included in both the translational and rotational dynamics.
Gaussian processes are also applied to learn the residual physics with online fashion
\cite{torrenteDataDrivenMPCQuadrotors2021, tase_learnCtrl}.

Despite the effectiveness of the above approaches,
the learning process is usually done in a label-feature fashion,
where the labels are the state derivatives or control inputs.
Such labels are usually noisy in real-world applications
and sometimes require extra sensors and specialized filters to obtain.
In addition, the learning process is usually done with a single-step prediction,
which fails to guarantee a stable and accurate long-horizon prediction.
Learning with long-horizon prediction is studied in \cite{longhorizon2024},
where the model is trained to predict the future trajectory recursively with high accuracy.
However, the method requires a fixed integration step size, which lacks the feasibility
of prediction with arbitrary integration step sizes.

Neural ordinary differential equations (Neural ODEs) \cite{chenNeuralOrdinaryDifferential2019}
have been introduced to learn the dynamics of systems with continuous-time
differential equations from a long-horizon integration perspective.
It enables prediction with arbitrary integration step sizes
and avoids the direct usage of noisy state derivatives as labels.
A recent work \cite{KNODE_MPC} applies Neural ODEs to learn the residual physics and enjoys
a good long-horizon prediction performance, but relies on motor thrusts as inputs.
Our learning method has a similar idea to Neural ODEs
but utilizes a closed-looped hybrid model,
which helps reduce error accumulation over time while only requiring 
smooth state trajectories and reference trajectories for training.

\subsection{Controller Synthesis with Learning-based Model}
Embedding a learning-based model into controller synthesis is
non-trivial in existing works.
Some works focus on embedded data-driven dynamics into compensation \cite{jiaEVOLVEROnlineLearning2024}
with noticeable results in improving the tracking performance.
Adaptive neural network is also a promising approach
for controller synthesis \cite{JacobianUnknown_Lyu2020,richardsAdaptiveControlOrientedMetaLearningNonlinear2021,
oconnell2022neuralfly,CompositeDisturbance_AerialManip_2025},
where the controller is adapted to the residual physics with adaptive laws.
Such methods enable direct adaptation to external disturbances by pretraining
deep neural networks as basis functions.
However, they introduce a more complicated controller structure and require
more complicated tuning tasks.
Model predictive controllers can directly utilize
data-driven dynamics 
into the optimization problem
\cite{torrenteDataDrivenMPCQuadrotors2021, tase_nn_MPC, KNODE_MPC} 
but rely on heavy onboard computing for real-time applications 
\cite{salzmannRealtimeNeuralMPCDeep2023, savioloPhysicsInspiredTemporalLearning2022a}
since learning-based dynamics could be strongly nonlinear.
The methods mentioned above mainly focus on taming disturbances
at the cost of increasing the complexity of the controller.
Our approach remains a standard controller while optimizing
the reference trajectory where the residual effects can be minimized
based on learning-based models.

\subsection{Trajectory Optimization for Aggressive Quadrotor Motion}
Planning aggressive motions for quadrotors is demanded in various applications.
Trajectory optimizations can be carried out via optimal control
techniques with point-mass model \cite{romeroTimeOptimalOnlineReplanning2022} 
or nonlinear state-space models \cite{foehnTimeoptimalPlanningQuadrotor2021,zhouEfficientRobustTimeOptimal2023}.
Besides, by utilizing the differentiable flatness properties,
the motion can be described using polynomials as flat outputs 
\cite{mellingerMinimumSnapTrajectory2011, gaoOptimalTimeAllocation2018, sunFastUAVTrajectory2021},
where a relaxation of computation complexity becomes possible.
However, neither of the approaches guarantees an accurate performance on 
closed-loop trajectory tracking since only nominal models
are employed for problem-solving.
A trajectory relaxation method with a nominal closed-loop model
is introduced in \cite{yangtrace2024} for precise control.
In our approach, the tracking performance is enhanced by minimizing the residual physics
along the reference trajectory.

\section{Preliminaries}

\subsection{Quadrotor Dynamics}

\begin{figure}[htbp]
  \begin{center}
  \includegraphics[width=0.3\textwidth]{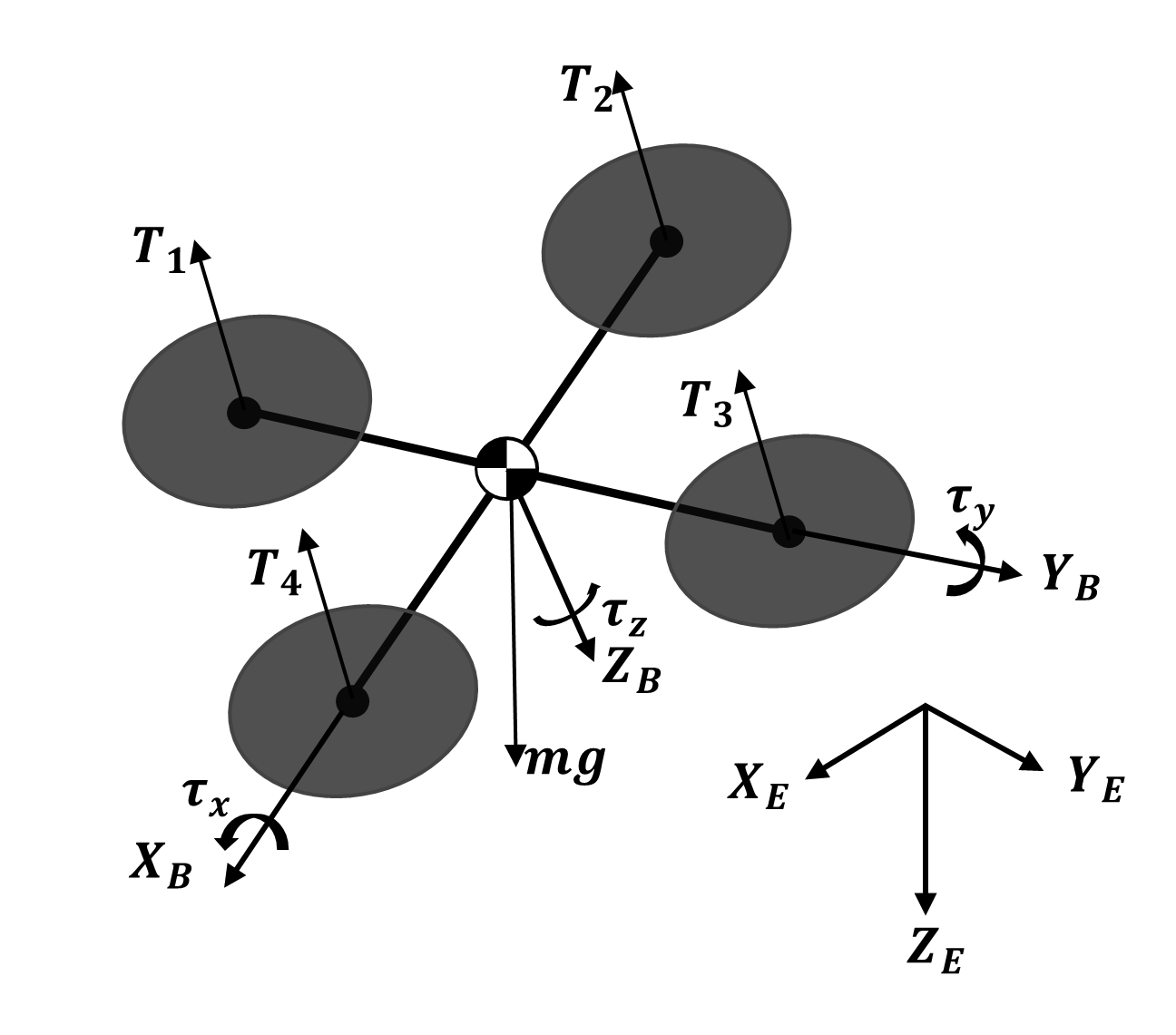}
  \end{center}
  \vspace{-1.0em}
  \caption{The schematic of the quadrotor system with the 
  definitions of the earth-fixed and the
  body-fixed frame.}
  \vspace{-1.0em}
  \label{quadrotor}
\end{figure}

As shown in Fig.\ref{quadrotor},
two coordinate systems are defined for the quadrotor:
the earth-fixed frame
$\bm{\mathcal{E}} = 
[\bm{X}_E, \bm{Y}_E, \bm{Z}_E]$ 
and the body-fixed
frame $\bm{\mathcal{B}} = [\bm{X}_B, \bm{Y}_B, \bm{Z}_B]$, where $\bm{Z}_E$ points to the ground.
Euler angles $\bm{\Theta}$ (roll-pitch-yaw) are used for attitude expression.
The whole quadrotor system is represented with a state vector 
$\bm{x} = [\bm{p}, \bm{v}, 
\bm{\Theta}, \bm{\omega}]^\top \in \mathbb{R}^{12}$ and an input vector
$\bm{u} = [T_1, T_2, T_3, T_4]^\top \in \mathbb{R}^4$ of motor thrusts.
The kinematics and dynamics are formalized as:
\begin{equation}
	\begin{aligned}
		&\dot {\bm p} = \bm v, \quad \dot {\bm v} = \bm a = 
		-\frac{1}{m}\bm Z_B T + g\bm Z_E\\
		& \dot {\bm{\Theta}} = \bm{W}(\bm{\Theta})\bm{\omega},\quad
		\bm J \dot {\bm \omega}= 
		-\bm \omega \times (\bm J \bm \omega) 
		+ \bm \tau\\
		& [T, \bm \tau]^\top = \bm{C} 
		[T_1, T_2, T_3, T_4]^\top
	\end{aligned}
\end{equation}
where $\bm{p}, \bm{v}$ are the
position and velocity in the earth-fixed frame and
$\bm{\omega}$ is the body rate defined
in the body-fixed frame, 
$\bm{W}(\cdot)$ refers to 
the rotational mapping matrix of Euler angle dynamics,
$T = T_1 + T_2 + T_3 + T_4$ is the total thrust,
$\bm{\tau}$ is the torque vector produced by the motors thrusts,
$\bm{C}$ is the control allocation matrix. For more details refer to \cite{JiaDragUtil2022}.

\subsection{Differential Flatness-Based Controller}
Differential Flatness-Based controller (DFBC) 
is a widely applied control scheme
and favored by tons of applications 
\cite{mellingerMinimumSnapTrajectory2011, faesslerDifferentialFlatnessQuadrotor2018,
morrellDifferentialFlatnessTransformations2018a}.
By receiving the flat outputs
$\bm{\bar \Psi} = [\bm{p}, \bm{v}, \bm{a}, \bm{j}]$,
i.e., the positional signal and its higher-order derivatives, as
the reference signal (yaw motion excluded), DFBC computes the desired motor thrusts
under the state feedback $\bm{x}$.

\subsection{Model Predictive Controller}
MPC works in the form of receding-horizon trajectory optimizations with a dynamic model and
then determines the current optimal control input. Approving optimization results highly rely on
accurate dynamical models. Befitting from the powerful representation capability of neural networks
for complex real-world physics, noticeable works 
\cite{torrenteDataDrivenMPCQuadrotors2021,tase_nn_MPC,savioloPhysicsInspiredTemporalLearning2022a,KNODE_MPC} 
have demonstrated that models incorporating first principles with learning-based
components can enhance control performance.

\section{Methodology}

\begin{figure}[htbp]
 \begin{center}
 \includegraphics[width=0.45\textwidth]{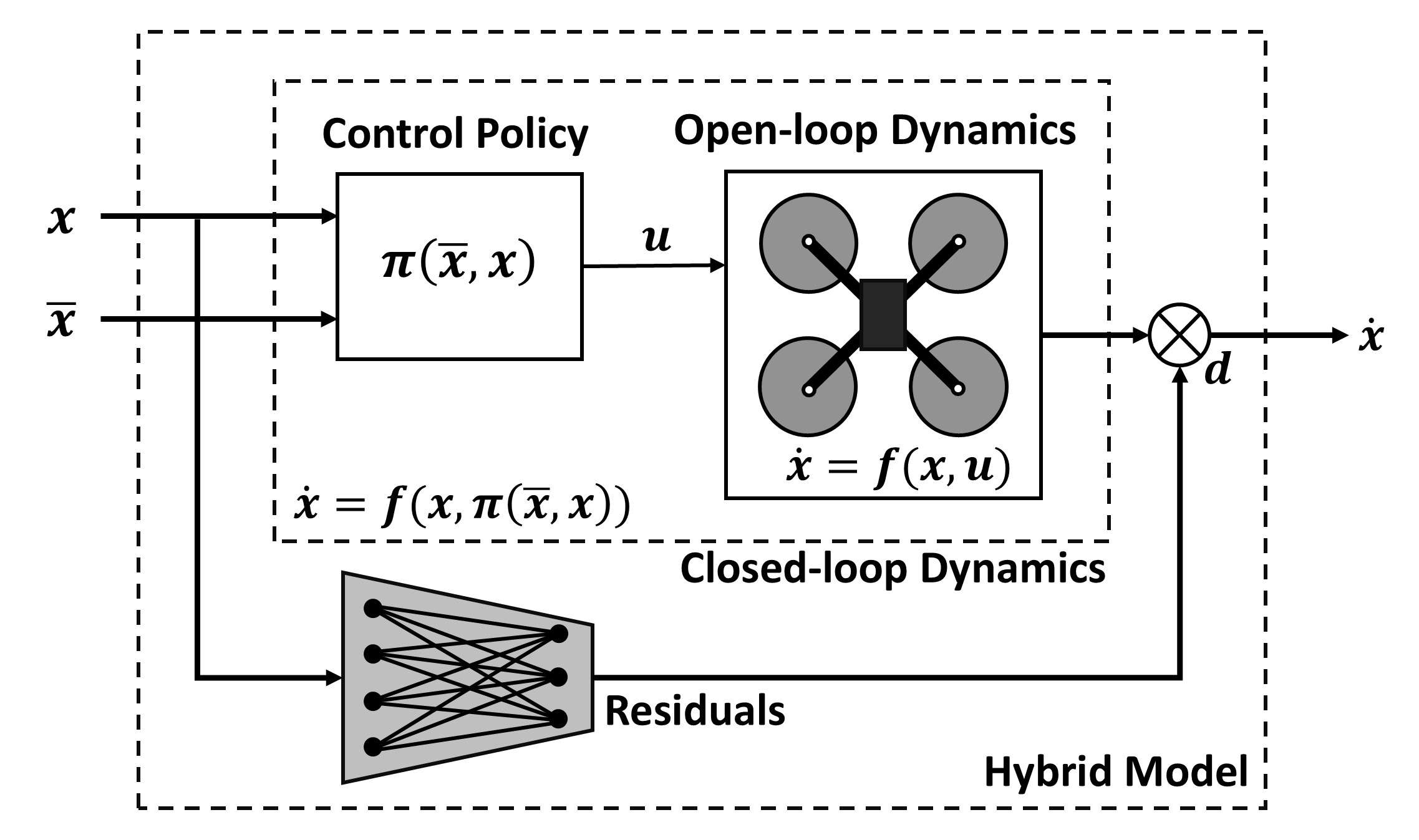}
 \end{center}
 \vspace{-1.0em}
 \caption{The proposed hybrid model. 
 Given an open-loop dynamics and a control policy,
 the closed-loop dynamics could be constructed 
 and augmented with a learning-based part
 to capture residual physics.}
 \vspace{-1.0em}
 \label{hybrid model}
\end{figure}

At first, a hybrid closed-loop dynamics with augmented residual physics is introduced.
Subsequently, a self-supervised residual learning algorithm is developed to capture the uncertainty only through state trajectories. 
Finally, using the learned residual model, 
a residual-minimal trajectory optimization algorithm is proposed, ending up with control-friendly trajectories.

\subsection{Hybrid Closed-loop Dynamics}
Given an open-loop dynamics 
$\bm{\dot x} = \bm{f}(\bm{x}, \bm{u})$ and 
a control policy $\bm{u} = \bm{\pi}(\bm{x}, \bm{\bar x})$,
the nominal closed-loop dynamics could be constructed as
$\bm{\dot x} = \bm{f}(\bm{x}, \bm{\pi}(\bm{x}, \bm{\bar{x}}))
= \bm{\phi}(\bm{x}, \bm{\bar x})$,
where $\bm{x} \in \mathbb{R}^n, \bm{\bar x} \in \mathbb{R}^r$, $\bm{u} \in \mathbb{R}^p$
represent the state, reference signal, and control input, respectively.
We study the case where the open-loop dynamics $\bm{f}(\cdot)$ is imperfect
with a state-related residual term $\bm{d}(\bm{x}) \in \mathbb{R}^n$:
\begin{equation}
 \bm{\dot x} = \bm{f}(\bm{x}, \bm{\pi}(\bm{x}, \bm{\bar{x}})) + \bm{d}(\bm{x})
 = \bm{\phi}(\bm{x}, \bm{\bar x}) + \bm{d}(\bm{x})
 \label{hybrid}
\end{equation}
where $\bm{d}(\bm{x})$ has the inputs of current state,
and it is defined as a learnable function parameterized by $\bm{\xi} \in \mathbb{R}^m$.
We note our hybrid closed-loop model (see Fig.\ref{hybrid model}) as $\bm{\dot x}
= \bm{\phi}(\bm{x}, \bm{\bar x}) + 
\bm{d}(\bm{x}, \bm{\xi}) 
= \bm{\Phi}(\bm{x}, \bm{\bar{x}}, \bm{\xi})$.

For quadrotor systems, residual physics can be
parametric uncertainties such as the bias on mass, inertia
and arm length, dynamic effects of the propeller-motor system,
and aerodynamic drag during high-speed flight.
Next, we apply a self-supervised learning algorithm to capture
the residual physics
$\bm{d}(\bm{x}, \bm{\xi})$.

\begin{remark}
  Using the closed-loop hybrid model for residual learning indicates that the neural network is learned under the effect of an existing controller.
  With the residual physics defined to be only state-related and sufficient control excitation provided by the reference trajectory,
  the residual physics identified from the closed-loop data generated by a specific controller can generalize well to other controllers,
  including optimization-based controllers such as MPC.
  Based on further tests, this holds well in quadrotor applications as long as the neural network is not overly complicated.
\end{remark}

\subsection{Self-Supervised Learning of Residuals}
\subsubsection{Problem Formulation}
Traditional learning methods require labels and features
for training using gradients computed via backpropagation.
However, the higher-order states in real-world systems are noisy labels.
Meanwhile, it also fails to guarantee a stable long-horizon prediction
due to the accumulation of learning errors in the state derivatives.
In this work, the challenge is addressed by differentiable training with integration using state trajectories.
Its gradients are computed by comparing the rollouts
of hybrid dynamics and the real-world closed-loop trajectories.

Firstly, we formulate the learning problem as an optimization problem:
\begin{equation}
 \begin{aligned}
  & \min_{\bm{\xi}} \sum_{k=1}^{N-1} \;l_k(\bm{x}_k, {\bm{x}_k}_r, \bm{\xi})
 + l_N(\bm{x}_N, {\bm{x}_N}_r)\\
  & s.t.\;\;\bm{x}_{k+1} = 
 \bm{\Phi}_{RK4}(\bm{x}_k, \bm{\bar{x}}_k, \bm{\xi})
 \end{aligned}
 \label{learningOC}
\end{equation}
where 
$\bm{x}_{k+1} = 
\bm{\Phi}_{RK4}(\bm{x}_k, \bm{\bar{x}}_k, \bm{\xi})$
refers to the discretized form of \eqref{hybrid} using 4$^{th}$ order Runge-Kutta method
with fixed discrete steps, 
${\bm{x}_k}_r$ denotes real-world recorded state trajectories, and
$l_k(\cdot)$, $l_N(\cdot)$ are defined to quantify 
the state differences between model rollout $\bm{x}_k$ 
and real-world state trajectory ${\bm{x}_{k}}_r$.
In this article, we select the functions in a weighted quadratic form: 
$({\bm{x}_{k}}_r-{\bm{x}_{k}}(\bm{\xi}))^\top \bm{L}_k
({\bm{x}_{k}}_r-{\bm{x}_{k}}(\bm{\xi}))$. 

\subsubsection{Gradient Computing}
The optimization problem is an unconstrained optimal control problem.
To obtain the optimal condition of the solution, the Lagrangian
of the problem is constructed:
\begin{equation}
  \mathcal{L} = J + \sum_{k=1}^{N-1} \bm{\lambda}_{k+1}^\top 
  (\bm{\Phi}_{RK4}(\bm{x}_k, \bm{\bar{x}}_k, \bm{\xi}) - \bm{x}_{k+1})
\end{equation}
where $\bm{\lambda}$ is the Lagrange multiplier of the dynamic constraints \eqref{rollout},
$J$ is the objective function in \eqref{learningOC}.
Subsequently, the first-order necessary condition for the constrained optimality is given as:
\begin{equation}
  \nabla_{\bm{x}} \mathcal{L} = \bm{0},
  \quad \nabla_{\bm{\lambda}} \mathcal{L} = \bm{0},
  \quad \nabla_{\bm{\xi}} \mathcal{L} = \bm{0}
\end{equation}
The above conditions can be further expanded, forming a set of equations for 
gradient computation. We reformed the conditions using Hamilton equation, which
is commonly used while solving optimal control problems.
\begin{align}
 & \begin{aligned}
 H &= J + 
  \sum_{k=1}^{N-1} \bm{\lambda}_k^\top
 \bm{\Phi}_{RK4}(\bm{x}_k, \bm{\bar{x}}_k, \bm{\xi})
 \end{aligned}\\
 &\bm{x}_{k+1} = \nabla_{\bm{\lambda}} H 
 = \bm{\Phi}_{RK4}(\bm{x}_k, \bm{\bar{x}}_k, \bm{\xi}),
 \;\bm{x}_1 = \bm{x}(0) \label{rollout}\\
 & \bm{\lambda}_k = \nabla_{\bm{x}}H 
 = \nabla_x l_k + (\frac{\partial \bm{\Phi}_{RK4}}{\partial{\bm{x}}})^\top
 \bm{\lambda}_{k+1} \label{adjoint},
 \;\bm{\lambda}_N = \frac{\partial{l_N}}{\partial{\bm{x}_N}}\\
 & \begin{aligned}
 \frac{\partial{H}}{\partial{\bm{\xi}}}
 = \sum_{k=1}^{N-1} \;\nabla_{\bm{\xi}} l_k + \bm{\lambda}_{k+1}^\top
  \nabla_{\bm{\xi}} \bm{\Phi}_{RK4} = 0
 \end{aligned} \label{gradient}
\end{align}
where $H$ stands for the Hamiltonian of this problem.
\eqref{rollout} and \eqref{adjoint} refer to the discrete dynamics of the nominal state
and the co-state or adjoint state, respectively.
Solving \eqref{gradient} can be done by applying gradient descent on $\bm{\xi}$.
The gradient is analytic and available
by sequentially doing forward rollout \eqref{rollout} of $\bm{x}$
and backward rollout \eqref{adjoint} of $\bm{\lambda}$,
where the latter one is also known as the term \textit{adjoint solve} 
or \textit{reverse-mode differentiation} \cite{chenNeuralOrdinaryDifferential2019}.
The whole gradient computation procedure is summarized in Algorithm \ref{gradient compute algorithm}.

\begin{algorithm}
  \caption{Analytic Gradient Computation}\label{gradient compute algorithm}
  \KwIn{Learning objective $l_k(\cdot), l_N(\cdot)$; 
        hybrid model $\bm{\Phi}_{RK4}$; 
        $\{\bm{x}_r, \bm{\bar x}\}$ trajectories}
  \KwOut{Gradient $\frac{\partial H}{\partial\bm{\xi}}$}
  $\bm{x} \leftarrow$
  Forward rollout of $\bm{\Phi}_{RK4}$ using \eqref{rollout}\;
  Compute $\nabla_{\bm{\theta}} l_k$, $\nabla_{\bm{x}} l_k$, $\nabla_{\bm{x}_N} l_N$, $\nabla_{\bm{x}} \bm{\Phi}_{RK4}$, $\nabla_{\bm{\theta}}\bm{\Phi}_{RK4}$;
  $\bm{\lambda} \leftarrow$
    Reverse rollout of $\nabla_{\bm{x}}H$ using \eqref{adjoint}\;
  $\frac{\partial H}{\partial\bm{\xi}} \leftarrow$ 
  Compute gradient using \eqref{gradient}\;
\end{algorithm}

\subsubsection{Learning Techniques}
The gradient computing only supports a single continuous state trajectory
and the computational complexity scales linearly with the trajectory length.
However, in real-world applications, multiple trajectory segments
with long horizons might be produced.
We introduce mini-batching as well as stochastic optimization methods
to deal with the drawbacks,
as summarized in Algorithm \ref{learning algorithm}.
The trajectory data $\mathcal{D}_{traj}$ is divided into $M$ segments,
each with a segment size of $N$.
With a given mini-batch size $s$, the gradient is updated with $s$ segments
at a time.
The update of the $\bm{\xi}$ could be done using \textit{ADAM} \cite{kingmaAdamMethodStochastic2017} 
or other stochastic optimizers.
The whole learning algorithm is implemented using CasADi \cite{Andersson2019}.
\begin{algorithm}
 \caption{Batch Self-Supervised Learning}\label{learning algorithm}
 \KwData{$\{\bm{x}_r, \bm{\bar x}\}$ trajectories $\mathcal{D}_{traj}$}
 \KwIn{learning objective $l_k(\cdot), l_N(\cdot)$; 
 hybrid model $\bm{\Phi}_{RK4}$;
 mini-batch size $s$}
 \KwOut{Learning parameters $\bm{\xi}$}
 $i\leftarrow 0$\;
 $\bm{\xi}\leftarrow Initialize$\;
 $\{{\bm{X}_r}_{1:M}, \bm{\bar X}_{1:M}\}_{1:s} 
 \leftarrow \text{Minibatching}(\mathcal{D}_{traj}, N, s)$\;
 \While{$i < i_{max}$}
 {
  \For{j = $1, \cdots, s$}
  {
    $\text{grad} \leftarrow \bm{0}$\;
    \For{$\{{\bm{x}_r}_{1:N}, \bm{\bar x}_{1:N}\}$ 
    in $\{{\bm{X}_r}_{1:M}, \bm{\bar X}_{1:M}\}_j$}
    {
    $\frac{\partial H}{\partial\bm{\xi}} \leftarrow$
    AnalyticGradientComputation \ref{gradient compute algorithm}\;
    $\text{grad} \leftarrow \text{grad} + \frac{\partial H}{\partial\bm{\xi}}$\;
    }
    $\bm{\xi} \leftarrow $Optimizer($\bm{\xi}, \text{grad}$)\;
  }
  $i\leftarrow i+1$\;
 }
\end{algorithm}

\begin{remark}
  Controllers with embedded integration 
  (e.g. a DFBC controller with Proportional–Integral–Derivative setup) can be denoted as 
  $[\bm{\dot x}, \bm{\dot z}]^\top = \bm{\pi}(\bm{z}, \bm{x}, \bm{\bar x})$, 
  where $\bm{z}$ is the auxiliary state of the controller.
  In the proposed Algorithms \ref{learning algorithm},
  $[\bm{x}, \bm{z}]^\top$ turns out to be the state instead of $\bm{x}$ only.
\end{remark}

\subsection{Control-friendly Trajectory Optimization}
\subsubsection{Problem Formulation}
Once the hybrid model has been learned, it is possible
to utilize the residual model in trajectory optimization to achieve high control precision:
\begin{equation}
 \begin{aligned}
  & \min_{\bm{u}, t_f}\;
  \int_0^{t_f} \;\bm{d}_{\bm{\xi}}(\bm{x})^\top \bm{d}_{\bm{\xi}}(\bm{x}) + \lambda_r \bm{u}^\top \bm{u}\;dt\\
  &\begin{aligned}
 s.t.\;\;
  &\bm{\dot x}
 = \bm{f}(\bm{x}, \bm{u}) + \bm{d}_{\bm{\xi}}(\bm{x})\\
  &\bm{x} \in \mathbb{X},\; \bm{u} \in \mathbb{U}
 \end{aligned}
 \end{aligned}
 \label{raw trajopt}
\end{equation}
where $t_f$ refers to the final trajectory time, 
$\bm{d}_{\bm{\xi}}(\bm{x})$ is the learned residual model,
$\lambda_r$ is the regularization weight for the control input,
$\mathbb{X}, \mathbb{U}$ are feasible sets of state and control, respectively.
Distinguishing from classical trajectory optimization problems 
\cite{romeroTimeOptimalOnlineReplanning2022,
foehnTimeoptimalPlanningQuadrotor2021,
zhouEfficientRobustTimeOptimal2023, 
mellingerMinimumSnapTrajectory2011, 
gaoOptimalTimeAllocation2018,sunFastUAVTrajectory2021}, 
our approach focuses on minimizing the residual effects
that are not considered during controller synthesis,
via defining objective functions.

\subsubsection{Sparse Formulation}
Since the problem \eqref{raw trajopt} is in continuous form,
we apply a direct multiple shooting method to discretize the problem.
The idea is to divide the time horizon into $N$ segments
with a variable step size $h_k$,
and the state and control input are sampled at each segment.
Such transcription ends up with a nonlinear programming (NLP) problem:
\begin{equation}
 \begin{aligned}
  & \min_{{x}_k, {u}_k, h_k}\;
  \sum_{k=1}^{N-1} \;\bm{d}_{\bm{\xi}}(\bm{x}_k)^\top \bm{d}_{\bm{\xi}}(\bm{x}_k) + \lambda_r \bm{u}_k^\top \bm{u}_k\\
  &\begin{aligned}
 s.t.\;\;
  &\bm{x}_{k+1}
 = \bm{f}_{RK4}(\bm{x}_k, \bm{u}_k, h_k)\\
  &\bm{x}_0 = \bm{x}(0),\;\bm{x}_N = \bm{x}(N)\\
  &\bm{u}_{lb} \leq \bm{u}_k \leq \bm{u}_{ub}\\
  &h_{lb} \leq h_k \leq h_{ub}\\
  &g(\bm{x}) \leq 0
 \end{aligned}
 \end{aligned}
 \label{NLP trajopt}
\end{equation}
where $N$ is the number of discrete nodes,
$\bm{u}_{lb}, \bm{u}_{ub}$ are the lower and upper bounds of the control input,
$g(\bm{x}) \leq 0$ refers to state-related constraints,
such as obstacle avoidance or waypoint passing, 
$h_k$ is the step size of the numerical integration,
and $h_{lb}, h_{ub}$ are the lower and upper bounds of the step size.
By optimizing $h_k$ with a box constraint, one can achieve flexible final time
with feasible numerical integration.
Favored by NLP solvers, 
direct multiple shooting takes the integration terms $\bm{x}, h$
as decision variables to provide a sparse structure of the Jacobians and Hessians.
We apply the CasADi \cite{Andersson2019} toolkit with the Ipopt solver to formulate and solve \eqref{NLP trajopt}.

\subsubsection{Warm-starting}
Since the hybrid model might come with strongly nonlinear learning parts,
a good initial guess would facilitate the NLP solver.
We obtain the initial guess of state $\bm{x}_k$ by linearly interpolating
the positions between waypoints with other dimensional remained zeros.
The initial guess of control input $\bm{u}_k$ is obtained using the
thrust at hovering, i.e., $\bm{u}_k = \frac{mg}{4}$.
The initial guess of step size $h_k$ is set to a constant value, e.g., $0.01s$.

\subsubsection{Constraints on the Reference Trajectory}
Since the trajectory optimization problem \eqref{raw trajopt} outputs
a reference trajectory $\bm{x}, \bm{u}$
for control applications,
we need to ensure that the reference trajectory
$\bm{x}$ is feasible for the nominal closed-loop system.
For model predictive control (MPC) applications,
considering augmented open-loop dynamics for trajectory optimization
could favor the tracking performance since residual
effects are considered.
However, for differential flatness-based controllers, 
the reference trajectory $\bm{\bar x}$ should be in the form of
nominal open-loop dynamics since a DFBC is designed to track
a reference trajectory with differential flatness properties.
Augmentation with the learned residual model could break the flatness properties
of the reference trajectory, which is not desired.

\subsubsection{Computational Complexity} \label{sec:computation}
The computational complexity of the proposed trajectory optimization
problem \eqref{NLP trajopt} is mainly determined by the number of decision variables
and the optimization solver.
The proposed method shares the same structure as the classical trajectory optimization
problem transcript by direct multiple shooting.
The total number of decision variables is $N \cdot (n + p + 1)$,
where $n$ is the dimension of the state vector $\bm{x}$, 
$p$ is the dimension of the control input $\bm{u}$, and $N$ is the number of segments.
Additionally, we have $N$ continuous constraints and $N \cdot (n + p + 1)$ box constraints,
skipping the waypoint constraints for simplicity.
Thus, the Karush-Kuhn-Tucker (KKT) system of the NLP problem has 
the size of $S = 2N \cdot (n + p) + 3N$.
The computational complexity of the NLP solver can be given as $\mathcal{O}(k \cdot S^3)$,
where $k$ is the number of iterations for convergence.
By utilizing the sparsity of the Jacobian and Hessian matrices,
the complexity can be reduced.
However, the cost of Jacobian and Hessian computation with a Neural Network (NN) model
is still high, we provide a benchmark comparison in the simulation section for further discussion.

\begin{remark}
    While small $\lambda_r$ encourages the optimizer to find a trajectory
    with minimal residual effects, it can introduce difficulties into the optimization,
    especially when the learned residual model is strongly nonlinear.
\end{remark}

\begin{figure*}[htbp]
  \centering
  \includegraphics[width=1.0\textwidth]{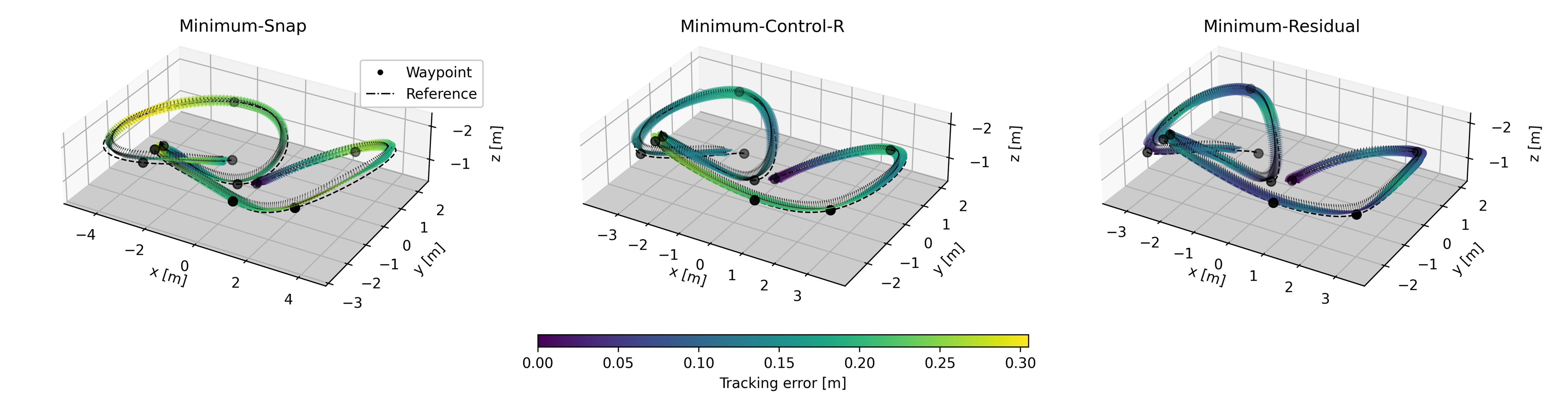}
  \caption{Trajectory tracking results of the proposed trajectory optimization 
 and the compared methods.
 Using a nominal MPC for trajectory tracking,
 the minimum-residual trajectory achieves the best tracking performance among others.
 The mean aerodynamic drag of the minimum-residual trajectory is $0.817N$, which
 is significantly lower than the $1.845N$ that appeared in the minimum-snap trajectory and
  $1.483N$ in the minimum-control-R trajectory.}
  \vspace{-1.0em}
  \label{sim_trajopt_compare}
\end{figure*}

\section{Simulations}

\subsection{Learning Aerodynamic Residuals}
In the simulation, the efficiency of the proposed learning 
framework to capture the aerodynamic residuals of a quadrotor is demonstrated.
Several works \cite{faesslerDifferentialFlatnessQuadrotor2018,bauersfeldNeuroBEMHybridAerodynamic2021,JiaDragUtil2022}
have been done on the aerodynamic modeling of a quadrotor system
and achieved extraordinary progress.
A classical model \cite{faesslerDifferentialFlatnessQuadrotor2018}
is deployed for aerodynamic simulation, 
in which the aerodynamic drag can be written as:
$\bm{R}\bm{D}\bm{R}^\top \bm{v}$,
where $\bm{R}$ refers to the current rotational matrix that maps
the body-fixed frame $\bm{\mathcal{B}}$ 
to the earth-fixed frame $\bm{\mathcal{E}}$, 
and $\bm{D}$ is a coefficient matrix.

\begin{figure}[htbp]
  \begin{center}
  \includegraphics[width=0.4\textwidth]{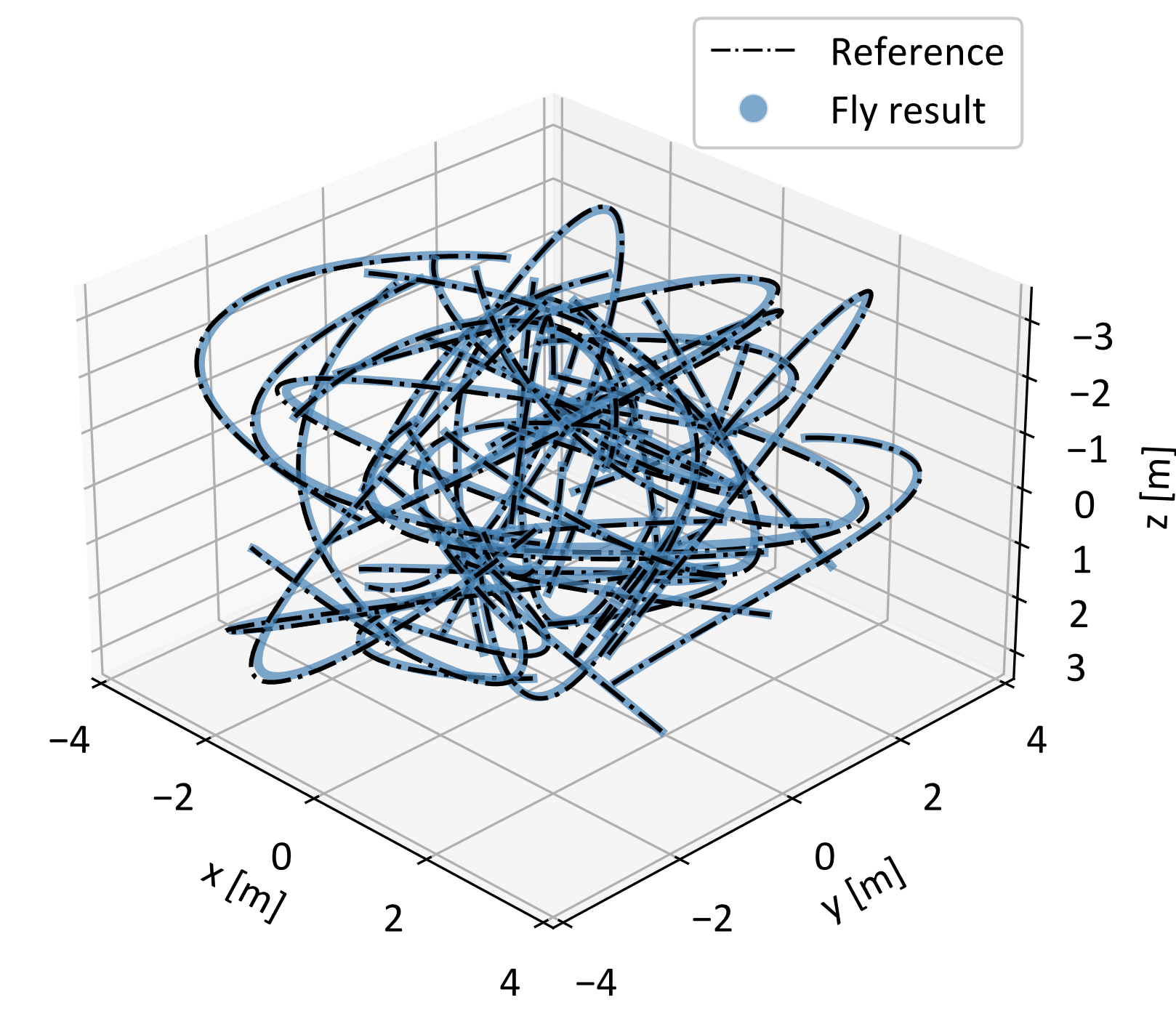}
  \end{center}
  \vspace{-1.0em}
  \caption{Trajectories generated with random waypoints for aerodynamic drag learning.
 Each trajectory contains 5 random waypoints and 200 discrete nodes with a step size of $0.02 \sec$.}
  \vspace{-1.0em}
  \label{data_sample}
 \end{figure}

\begin{figure}[htbp]
 \begin{center}
 \includegraphics[width=0.4\textwidth]{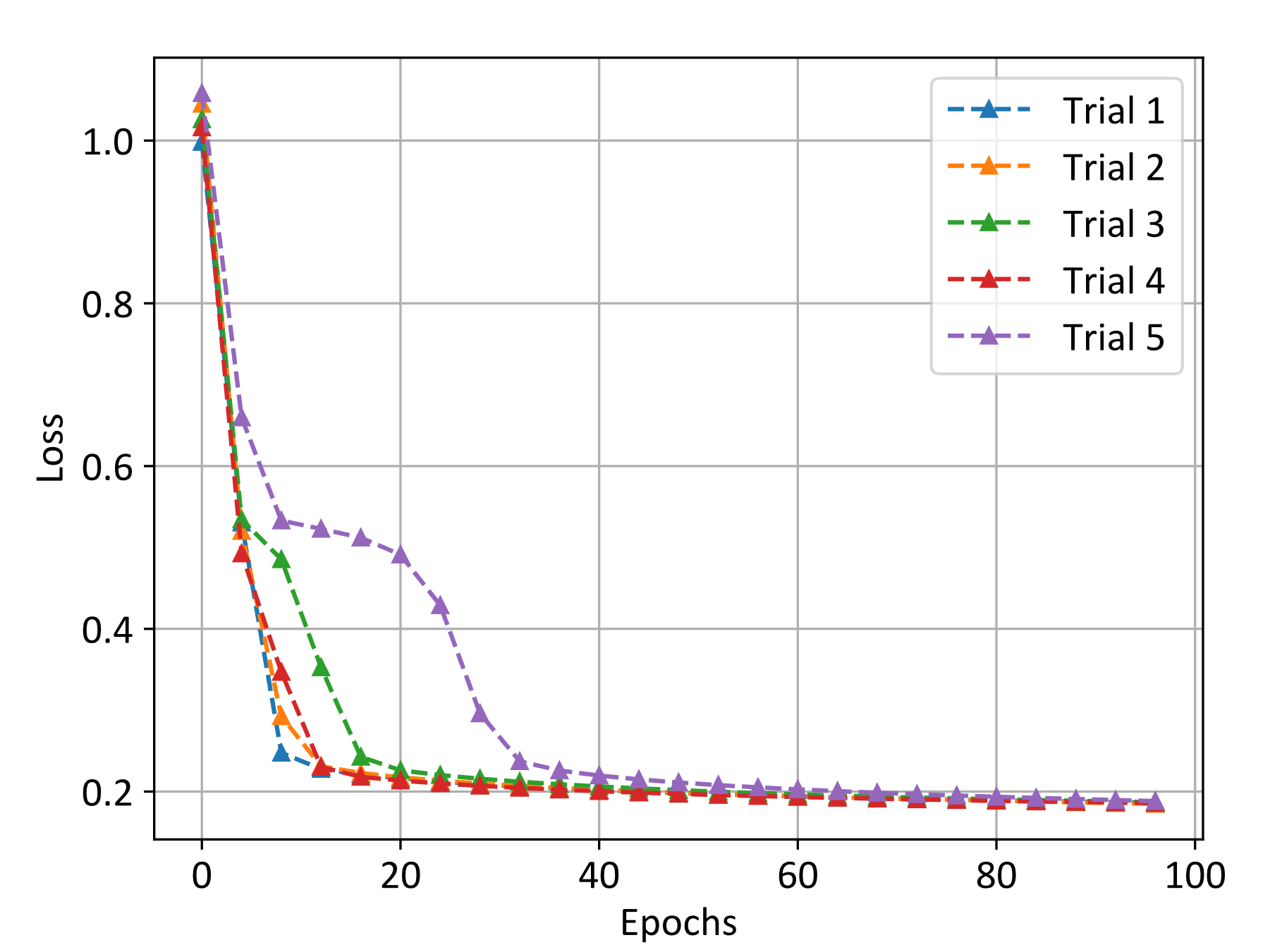}
 \end{center}
 \vspace{-1.0em}
 \caption{Learning curves of 5 trials with random initial guesses
 around zero initialization. 
 At around epoch 100, all trials achieve promising regression results.}
 \vspace{-1.0em}
 \label{sim_learningcurve}
\end{figure}

\begin{figure*}[htbp]
  \begin{center}
  \includegraphics[width=0.98\textwidth]{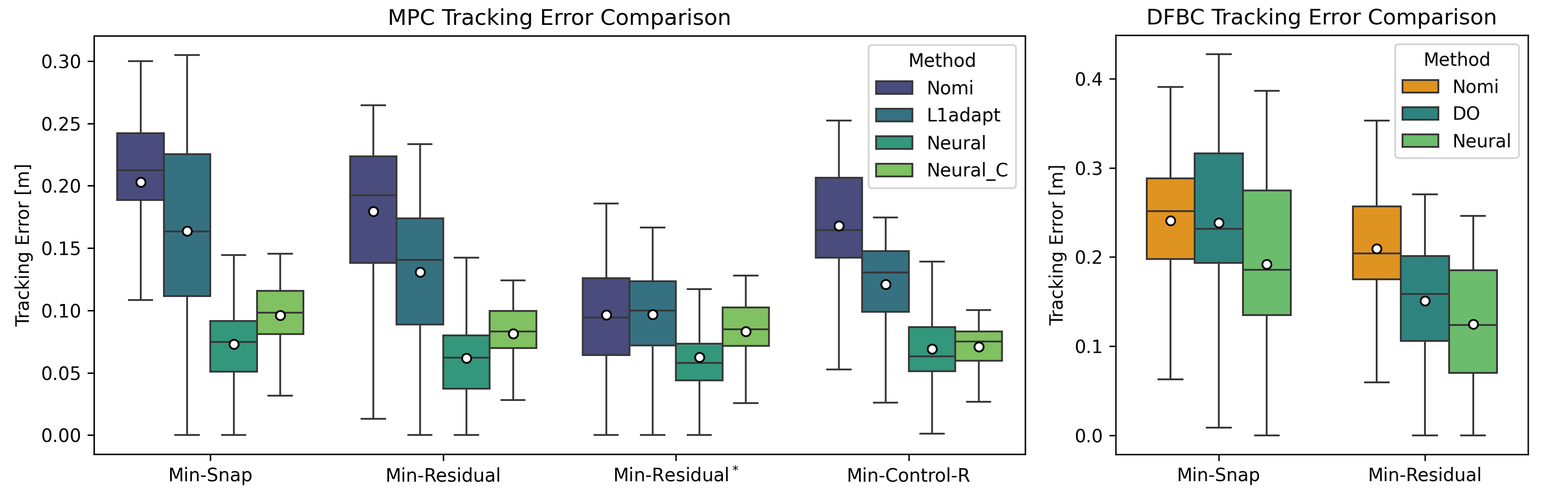}
  \end{center}
  \caption{
 Positional tracking error of the MPC and DFBC controllers with different trajectories.
 }
  \vspace{-1.0em}
  \label{sim_trackerr_boxplot}
\end{figure*}

\subsubsection{Data Generation}
40 trajectories are generated for learning by random sampling
the positional waypoints in a limited space, followed by optimizing polynomials that connect these
waypoints, as shown in Fig.\ref{data_sample}. 3 trajectories are selected for the validation of the
learned model. Around 8000 discrete states are sampled from the generated trajectories.
We split the dataset into a training set and a test set with a ratio of 8:2.

\subsubsection{Learning Setups}
To validate the learning framework, we directly learn the
aerodynamic drag from trajectory data.
The nominal model is augmented with a neural ODE describing the residual state derivatives $\bm d(\cdot)$ caused by the aerodynamic drag.
A multilayer perceptron (MLP) with two hidden layers of 32 units 
takes a concatenated input of the 6 nominal states $[v_x, v_y, v_z, \Theta_x, \Theta_y, \Theta_z]^\top$ 
and outputs the 3-dimensional aerodynamic drag.
The given trajectory data requires mini-batching
for gradient computation. The segment size $N$ is set to 50
while the mini-batch size $s$ is set to 10.
\textit{ADAM} is selected for stochastic optimization
with an initial learning rate of $10^{-2}$.
Lastly, the quadratic matrix $\bm{L}$ in the learning loss \eqref{learningOC} is 
defined as identical.

\subsubsection{Results}
The learning process of 5 trials with a random initial guess
are shown in Fig.\ref{sim_learningcurve}, the \textit{ADAM} algorithm
convergences with promising results. The root-mean-square error (RMSE) of
acceleration predicted by the hybrid model 
is $\bm{0.260}$ and $\bm{0.173}$ in training and testing trajectories,
compared with 0.684 and 0.603 provided using the nominal model only.

\subsection{Optimizing Trajectories for High-speed Flight}
The hybrid model with the residual physics captured can be further used 
for trajectory optimization for better control performance.
We show that the simulated aerodynamic residual could cause
a degradation in tracking accuracy.
The obtained control-friendly trajectory can improve the positional tracking performance in high-speed flight.
Here we compared our method with several trajectory optimization methods, showing that the proposed method
can achieve a better tracking performance with a standard controller and even achieve the tracking accuracy
of the state-of-the-art control methods.

\subsubsection{Trajectory Optimization Setups}
We apply a sequence of waypoints to generate trajectories of $10 \sec$ 
with discrete nodes $N$ of 400 and discrete step size of $0.025 \sec$.
The control input regularization weight $\lambda_r$ is set to $0.1$.
A minimum-snap trajectory is set for the baseline comparison.
Besides, we also compare the proposed method with a classical minimum-control 
effort trajectory optimization method
as well as a dynamic-feasible trajectory optimization method 
\cite{bansalLearningQuadrotorDynamics2016}.
The trajectories and their notations are summarized below:
\begin{itemize}
\item  \textit{Min-Snap}: Minimum-snap trajectory optimized using a discrete polynomial-based method \cite{mellingerMinimumSnapTrajectory2011,yangtrace2024}.
\item  \textit{Min-Residual}: Minimum-residual trajectory optimized with nominal dynamics.
\item  \textit{Min-Residual}$^*$: Minimum-residual trajectory optimized with augmented dynamics.
\item  \textit{Min-Control-R}: Minimum-control effort trajectory optimized with nominal dynamics, followed by
 a dynamic feasibility refinement that solves an optimal tracking problem with augmented dynamics \cite{bansalLearningQuadrotorDynamics2016}.
\end{itemize}

\subsubsection{Controller Setups}
For baseline comparison, 
we employ both a model predictive controller (MPC) 
and a differential flatness-based controller (DFBC) for trajectory tracking. 
Additionally, an $\mathcal{L}_1$-adaptive estimator 
and a disturbance observer (DO) is implemented to estimate aerodynamic drag, 
enabling feedforward compensation for the respective controllers.  
Finally, the learned neural ODE model is integrated into both controllers, 
resulting in state-of-the-art learning-based control approaches.
The controllers with neural integration are supposed to achieve better tracking performance
by compensating for the residual physics.
The controllers and their notations are summarized as follows:
\begin{itemize}
\item \textit{Nomi}: Baseline controller with nominal dynamics.
\item \textit{DO}: Feedforward compensation with disturbance observer \cite{guo_antidisturbance_2014} for DFBC.
\item \textit{$L_1$adapt}: Feedforward compensation with $\mathcal{L}_1$-adaptive estimator \cite{l1_adapt_ctrl,huangDATTDeepAdaptive2023} for MPC,
 the feedforward compensation is realized by applying a constant feedforward term within the receding horizon.
\item \textit{Neural}: For DFBC, it works as a feedforward compensation,
 where the learned residual model is integrated into disturbance estimation. 
 For MPC, the residual-augmented dynamics
 is used to predict the future trajectory \cite{KNODE_MPC}.
\item \textit{Neural-C}: An MPC approach similar to \cite{salzmannRealtimeNeuralMPCDeep2023} 
  with a classical neural network model that directly predicts the residual acceleration
 given the current state and control input. Such a model is trained with direct supervised learning, using ground-truth labeled drag data.
\end{itemize}
The MPC is implemented with a receding horizon of $0.5 \sec$ and a sampling time of $0.025 \sec$.
The DFBC is implemented with a PD translational controller and a PID body rate controller.

\subsubsection{Results}
As illustrated in Fig.\ref{sim_trajopt_compare},
the minimum-residual objectives lead to a very different result from the
minimum-snap and minimum-control ones.
With both trajectories of the same final time, 
it could be observed that the
minimum-residual trajectory tracking ends up with
a higher tracking accuracy as it passes through all waypoints,
indicating its control-friendly property.
By minimizing the aerodynamic effects, precise motion control
could also be achieved via a standard controller
without considering the aerodynamics during the synthesis process.

From Fig.\ref{sim_trackerr_boxplot}, the neural MPC
outperforms the nominal MPC and $\mathcal{L}_1$-adaptive MPC
in terms of positional tracking error for all trajectory candidates.
However, tracking control along $\textit{min-residual}^*$
leads to outstanding but similar performances for all tracking controllers.
This indicates that the proposed method can achieve
better tracking performance with a standard controller
and even achieve the tracking accuracy of the state-of-the-art control methods.
As a result, the design and parameter tuning of the data-driven controller
can be activated, which is beneficial for real-world applications.
Besides, the comparison between $\textit{min-residual}$ and $\textit{min-residual}^*$
shows the importance of considering the residual physics inside
the continuous constraints.
Similarly, the augmentation of neural feedforward helps the DFBC to achieve a better tracking performance.
However, the proposed trajectory brings a general improvement in
all control methods.

Lastly, through all MPC tracking control cases, $\textit{Neural}$ ends up with the lowest mean tracking error
and outperforms $\textit{Neural-C}$. Such results indicate the advantage of utilizing the self-supervised learning framework
with long-horizon integration loss compared with direct supervised learning with single-step labeled data.

\subsubsection{Ablation Study on Computational Complexity}
\begin{figure}[htbp]
  \centering
  \includegraphics[width=0.45\textwidth]{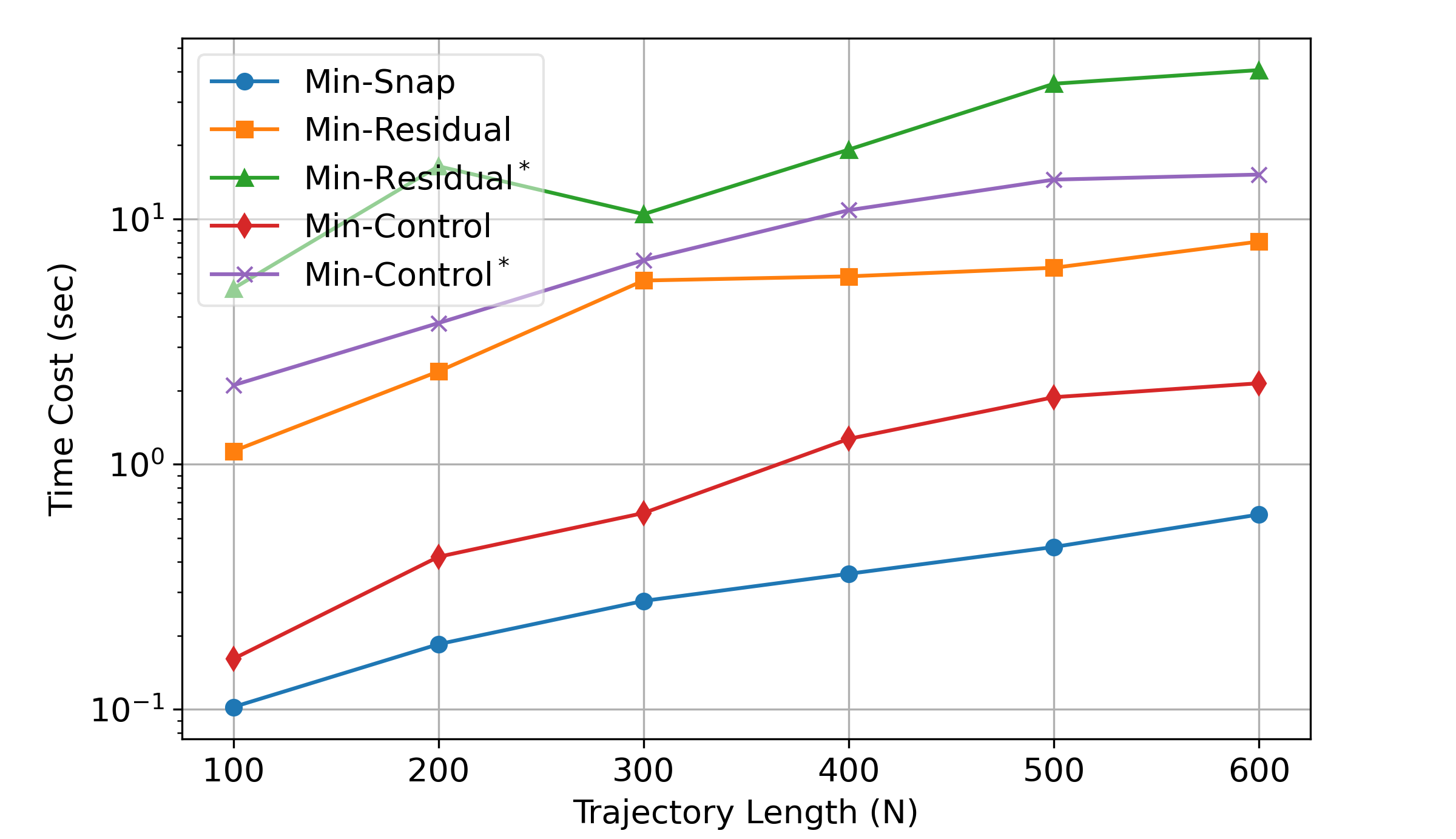}
  \caption{Comparison on the time cost of different trajectory optimization
 problems using the interior-point solver of Ipopt.
 }
  \label{benchmark_time}
  \vspace{-1.0em}
\end{figure}

Section \ref{sec:computation} shows that the proposed method
shares a similar computational complexity with classical
trajectory optimization methods.
However, the effect of the neural model could increase the number
of iterations to obtain a solution.
We test the computation cost of the proposed method
and the compared methods using on a laptop with an Intel i7-10750H CPU
and 16GB RAM.
As illustrated in Fig.\ref{benchmark_time}, the time cost of the proposed method
is comparable with that of the minimum-snap trajectory optimization problem.
This brings a promising result that the proposed method
can be applied to real-time applications with onboard computers, 
but might require further optimization to increase the computational efficiency.
A local convexification method \cite{salzmannRealtimeNeuralMPCDeep2023} could be applied to further reduce the time cost.
The \textit{Min-Control} method refers to a classical minimum-control effort trajectory optimization method
with nominal dynamics, while \textit{Min-Control$^*$} utilizes the augmented dynamics.
By comparing the time cost of the two methods, the effect of using neural dynamics
inside the constraints can be observed.

\begin{figure*}[htbp]
  \centering
  \includegraphics[width=0.95\textwidth]{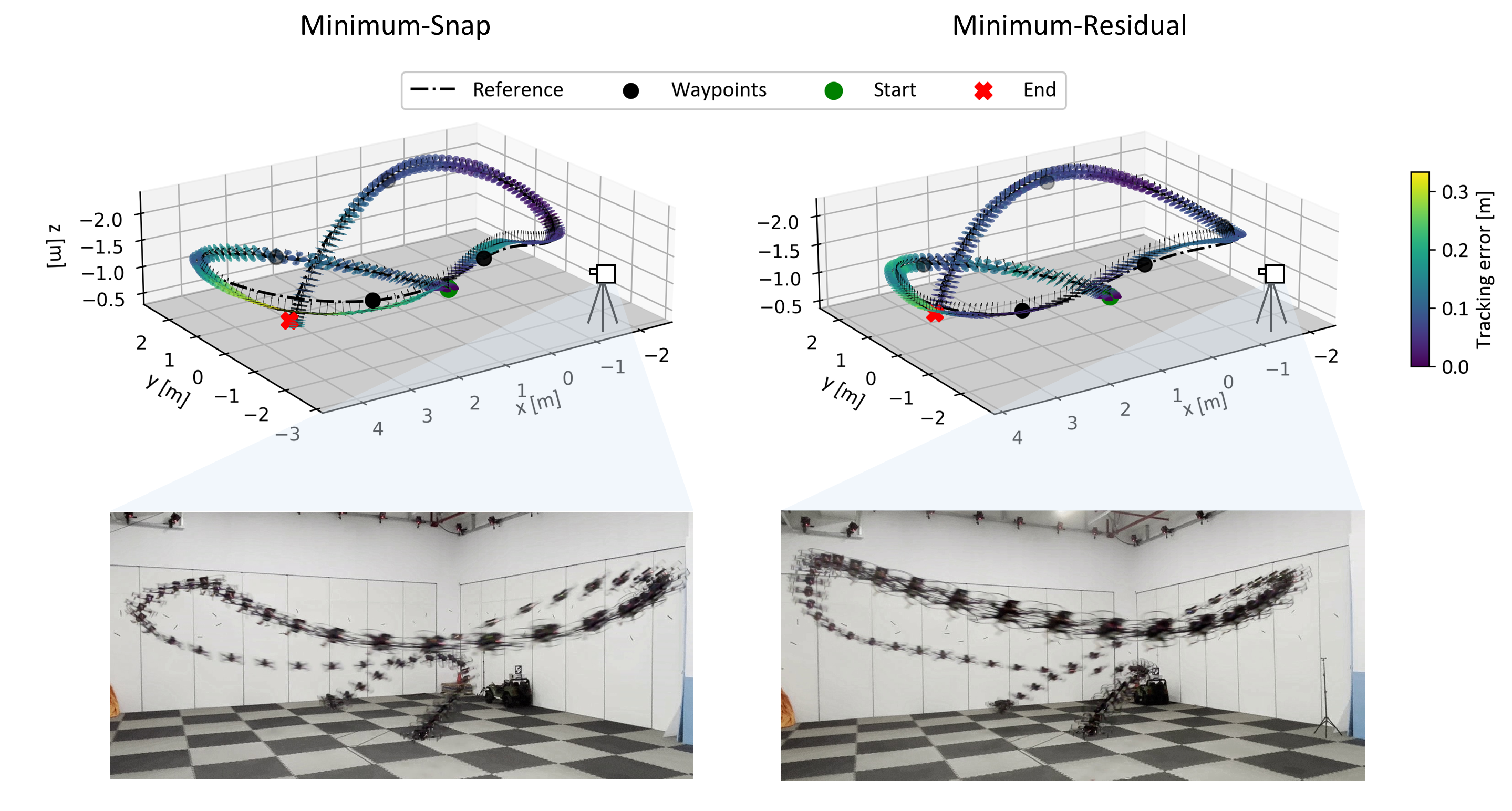}
  \caption{Trajectories optimized using minimum-snap and minimum-residual objectives
 together with the camera view of the real-world flight. 
 Tracking control with a DFBC baseline controller 
 on the minimum-residual trajectory ends up with
 lower positional tracking error.
 }
  \vspace{-1.0em}
  \label{experiment_sum}
\end{figure*}

\section{Real-world Experiments}
The presented framework is applied to a quadrotor platform to learn
real-world residual physics which is far more complex than that of the simulation.
Utilizing the hybrid model, the following minimum-residual optimization outputs aggressive
but control-friendly motions.
Real-world flight tests show a low trajectory tracking error while executing such motions.

\subsection{Hardware Setup}
Our experiment platform is equipped with an STM32-F7 microcontroller 
running onboard for the flight control algorithm.
Besides, an STM32-F4 microcontroller is set up for data recording, reception
and transmission with an ultra-wideband (UWB) telemetry module.
A motion capture system with 32 cameras running on the ground station
is used for the localization of the quadrotor.

\subsection{Learning Real-world Residual Dynamics}

\subsubsection{Data Generation} \label{sec:real_data_gen}
We generate four aggressive $\bm{\bar \Psi}$ trajectories, 
each approximately 15 seconds long, 
by randomly sampling 20 positional waypoints within a constrained space. 
These waypoints are connected using optimized polynomials. 
To augment the trajectory dataset with less aggressive variants, 
each trajectory is relaxed by applying four different decay rates to the trajectory time. 
After discretizing the trajectories with a time step of 0.02 seconds, 
we obtain over 10,000 discrete states and reference points in total.
We split the dataset into a training set and a test set with a ratio of 8:2.
To validate the generalizability of the hybrid model, 
we further generate another 3 trajectories by sampling various
waypoints and optimizing minimum-snap polynomials, but with a higher mean acceleration
that results in more aggressive motions that are \textit{out-of-distribution}.
This can be achieved by adjusting the time decay rate to be larger than 1.0.
The statistics of the generated datasets are summarized in Table \ref{tab:data_stat}.

\begin{table}[htbp]
  \centering
  \caption{Statistics of the Generated Datasets}
  \renewcommand\arraystretch{1.2}
  \label{tab:data_stat}
  \scalebox{0.95}{
    \begin{tabular}{|l|c|c|}
      \hline
      Metric & Training \& Test & \textit{Out-of-distribution} \\
      \hline
      Num. of Traj. & 16 & 3 \\
      Mean/Std. Vel. (m/s) & 1.595 / 0.854 & 2.721 / 1.350 \\
      Mean/Std. Acc. (m/s$^2$) & 2.478 / 1.971 & 4.403 / 2.437 \\
      Mean/Std. Jerk (m/s$^3$) & 4.726 / 5.300 & 8.845 / 5.732 \\
      \hline
    \end{tabular}
  }
  \vspace{-0.5em}
\end{table}

\subsubsection{Learning Setups}
The nominal model is augmented with a neural ODE that 
describes the residual state derivatives $\bm d(\cdot)$ caused by real-world physics.
We choose to model the residual physics in the acceleration space.
A multilayer perceptron (MLP) with 3 hidden layers of 32 units 
takes a concatenated input of the 6 nominal states $[v_x, v_y, v_z, \Theta_x, \Theta_y, \Theta_z]^\top$ 
and outputs the extra terms $[d_{v_x}, d_{v_y}, d_{v_z}]^\top$.
\textit{ADAM} is selected for stochastic optimization
with an initial learning rate of $10^{-3}$.
A mini-batch size of $s=16$ is set for the stochastic optimization.
The segment size $N$ is set to $25$.
Lastly, the quadratic matrix $\bm{L}$ in the learning loss is 
defined to be diagonal: 
$\bm{L} = diag\{\bm{1}_{2\times1},\;0,\;\bm{0.1}_{2\times1},\;0,\;\bm{0.1}_{2\times1},\;\bm{0}_{4\times1}\}$.
The convergence of the learning process appears at around 40 epochs, indicating
promising results of the learning setups.

\begin{remark}
  Only the residual physics in the translational acceleration is modelled, as the rotational dynamics of a quadrotor 
  is usually well-modeled with the nominal rigid-body dynamics. 
  The residual effects in the rotational dynamics could also be modeled with a similar approach, 
  but might be complicated to capture and generalize due to the high frequency of rotational motions.
\end{remark}

\subsection{Model Evaluation}
The hybrid model can be evaluated via rollouts with different reference trajectories and
comparing the predicted states with the truth states.
We first evaluate the model on the training set and
test set mentioned in Section \ref{sec:real_data_gen}.
Table \ref{tab:model_eval} shows the performance of both hybrid
and nominal models.
The results show that the hybrid model can achieve a lower RMSE
on both training and test sets, indicating that the hybrid model
can capture the residual physics of the real-world quadrotor.
The hybrid model is managed to narrow the gap between
the ideal nominal model and the real-world physics via the neural network. 
However, the learning errors are still slightly larger on the \textit{out-of-distribution} trajectories
compared with that of the training set and test set.
Future work could be focused on learning more accurate real-world
residuals by building a large dataset with more actuation and stimulation, 
as well as refining the structure of the hybrid model and
learning objectives.

\begin{table}[htbp]
  \centering
  \caption{Hybrid Model Evaluation on Real-world Data}
  \renewcommand\arraystretch{1.2}
  \label{tab:model_eval}
  \scalebox{0.95}{
    \begin{tabular}{|c|c|c|c|}
      \hline
      \multirow{2}{*}{Dataset} & \multirow{2}{*}{Metric} & \multicolumn{2}{c|}{Prediction Error RMSE} \\
      \cline{3-4}
      & & Nominal Model & Hybrid Model \\
      \hline
      \multirow{2}{*}{Training} 
        & Position (m) & 0.101 & \textbf{0.064} \\
        & Velocity (m/s) & 0.225 & \textbf{0.137} \\
      \hline
      \multirow{2}{*}{Test} 
        & Position (m) & 0.097 & \textbf{0.067} \\
        & Velocity (m/s) & 0.192 & \textbf{0.129} \\
      \hline
      \multirow{2}{*}{\textit{Out-of-distribution}}
        & Position (m) & 0.114 & \textbf{0.093} \\
        & Velocity (m/s) & 0.244 & \textbf{0.199} \\
      \hline
    \end{tabular}
  }
  \vspace{-1.0em}
\end{table}

\subsection{Aggressive Multi-waypoint Flight} \label{aggressive_flight}
\subsubsection{Trajectory Optimization Setup}
The trajectory is discretized with the $N=280$ with the step size of $0.03 \sec$,
sequentially connects 7-waypoints and forms a pretzel-shape trajectory.
The control input regularization weight $\lambda_r$ is set to $0.1$.
A minimum-snap trajectory is set for the baseline comparison.

\subsubsection{Controller Setup}
A DFBC is implemented with a PD translational controller and a PID body rate controller.
We further augment such baseline controller with a disturbance observer (DO)
and a neural feedforward compensation.

\subsubsection{Results}
\begin{figure}
  \centering
  \includegraphics[width=0.45\textwidth]{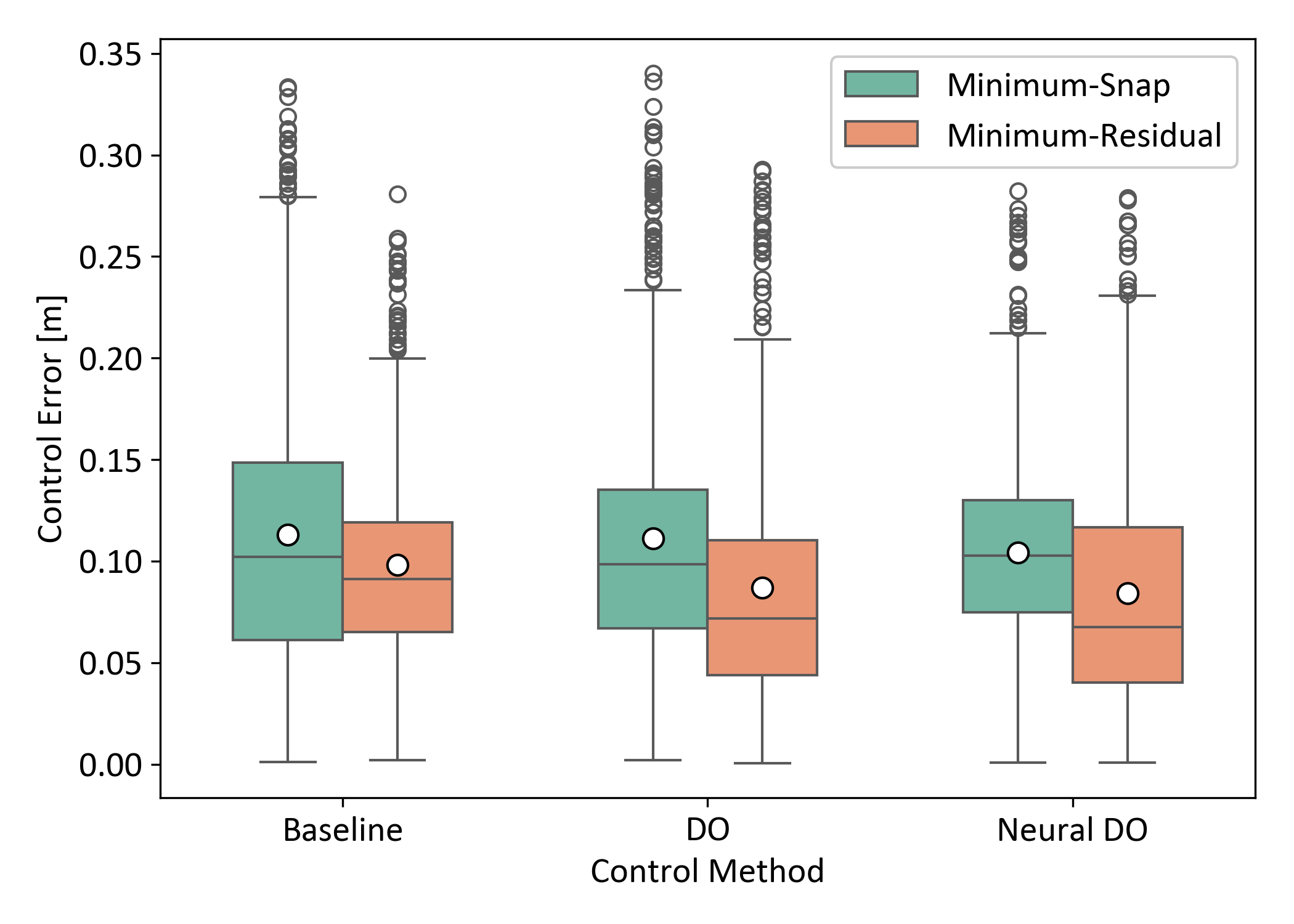}
  \caption{
  The real-world positional tracking error of the DFBC controllers with different trajectories.
 }
  \vspace{-1.0em}
  \label{exp_error_boxplot}
\end{figure}

Fig.\ref{experiment_sum} illustrates both the minimum-snap 
and minimum-residual trajectories optimized via
Algorithm \ref{NLP trajopt} together with the real-world trajectory
tracking motion.
From Table \ref{tab:exp_result} and Fig.\ref{exp_error_boxplot}, 
the minimum-residual trajectory ends up with less mean tracking error.
It is worth mentioning that using the baseline controller on the minimum-residual trajectory
outperforms using the state-of-the-art neural feedforward on the minimum-snap trajectory.
Such property indicates that the detailed design of the controller
can be alleviated, as the proposed method can achieve a better tracking performance
without considering the residual physics during the synthesis process.
The maximum jerk of the minimum-residual trajectory is larger 
than that of the minimum-snap trajectory.
This could occur as the reference trajectory optimization outputs higher jerks 
for residual physics rejection.
One possible outcome is that the residual physics is minimized 
but leads to a more aggressive motion, which might be less preferred in some applications.

\begin{table}[htbp]
  \centering
  \caption{Experimental Results for Aggressive Multi-waypoint Flight}
  \renewcommand\arraystretch{1.2}
  \label{tab:exp_result}
  \scalebox{0.95}{
    \begin{tabular}{|l|c|c|}
      \hline
      \textbf{Metric} & $\;\textbf{Min-Snap}\;$ & $\;\textbf{Min-Residual}\;$ \\
      \hline
        Max. Acceleration (m/s$^2$) & 9.392 & 9.091 \\
        Max. Jerk (m/s$^3$)         & 22.684 & 46.676 \\
        Mean Acceleration (m/s$^2$) & 4.338 & 3.933 \\
        Mean Jerk (m/s$^3$)         & 9.442 & 9.673 \\
        Mean Pos. Error \textit{Baseline} (m)     & 0.113 & \textbf{0.098} \\
        Mean Pos. Error \textit{DO} (m)    & 0.111 & \textbf{0.087} \\
        Mean Pos. Error \textit{Neural DO} (m)    & 0.104 & \textbf{0.084} \\
      \hline
    \end{tabular}
 }
  \vspace{-1.0em}
\end{table}

\section{Conclusion}
This letter introduces a framework that utilizes learning-based dynamics
in motion planning for higher control performance without fine-tuning or re-designing the controller.
The self-supervised learning algorithm captures the residual physics
of the nominal model, which is free of noisy labels
and enables precise prediction with long-horizon.
It further enables trajectory optimization that favors the control tasks.
As the residual physics unconsidered during controller synthesis
is minimized, trajectory tracking performance can be enhanced during the motion planning stage.
Finally, applications for planning aggressive but precise multi-waypoint flight of quadrotors,
are shown via content simulations and real-world experiments.
The trajectories are well-predicted by the learned real-world hybrid model with
stable long-horizon integration,
yet it still requires further refinement to improve the generalizability
for \textit{out-of-distribution} data.

\subsection*{A. Limitations and Future Work}
One of the primary drawbacks of the proposed framework is that
the learning-based method is offline and lack of online adaptation
mechanisms to tackle the external disturbances.
A method that enables online adaptation to the rapidly changed dynamic effects
while remaining the offline learned priors
would be favored in the further development.

Meanwhile, the minimization of the residual physics
does not necessarily lead to a mild trajectory (Table \ref{tab:exp_result}).
In some cases, the trajectory optimization problem
could lead to a more aggressive motion
that is less preferred in some applications.
It requires careful tuning of the regularization weight
to balance the trade-off between the residual physics and the control effort.

Another major drawback lies in the trajectory optimization method.
Neural networks with nonlinear
activation function could slow down the optimization
process, such an issue also appears in the application of real-time
model predictive control which uses neural networks as part of
its dynamics.
Future work might include exploring other categories of regressors
with less complexity compared with neural networks,
as well as alternatives of hybrid model structure that are simpler
and more interpretable.

\bibliographystyle{IEEEtran}
\bibliography{reference}

@ARTICLE{foehnTimeoptimalPlanningQuadrotor2021,
  author = {Philipp Foehn  and Angel Romero  and Davide Scaramuzza },
  title = {Time-optimal planning for quadrotor waypoint flight},
  journal = {Science Robotics},
  volume = {6},
  number = {56},
  pages = {eabh1221},
  year = {2021},
  doi = {10.1126/scirobotics.abh1221}}

@INPROCEEDINGS{gaoOptimalTimeAllocation2018,
  author={Gao, Fei and Wu, William and Pan, Jie and Zhou, Boyu and Shen, Shaojie},
  booktitle={2018 IEEE/RSJ International Conference on Intelligent Robots and Systems (IROS)}, 
  title={Optimal Time Allocation for Quadrotor Trajectory Generation}, 
  year={2018},
  volume={},
  number={},
  pages={4715-4722},
  doi={10.1109/IROS.2018.8593579}}

@ARTICLE{romeroModelPredictiveContouring2022,
  author={Romero, Angel and Sun, Sihao and Foehn, Philipp and Scaramuzza, Davide},
  journal={IEEE Transactions on Robotics}, 
  title={Model Predictive Contouring Control for Time-Optimal Quadrotor Flight}, 
  year={2022},
  volume={38},
  number={6},
  pages={3340-3356},
  doi={10.1109/TRO.2022.3173711}}

@ARTICLE{romeroTimeOptimalOnlineReplanning2022,
  author={Romero, Angel and Penicka, Robert and Scaramuzza, Davide},
  journal={IEEE Robotics and Automation Letters}, 
  title={Time-Optimal Online Replanning for Agile Quadrotor Flight}, 
  year={2022},
  volume={7},
  number={3},
  pages={7730-7737},
  doi={10.1109/LRA.2022.3185772}}

@ARTICLE{songReachingLimitAutonomous2023,
  author = {Yunlong Song  and Angel Romero  and Matthias Müller  and Vladlen Koltun  and Davide Scaramuzza },
  title = {Reaching the limit in autonomous racing: Optimal control versus reinforcement learning},
  journal = {Science Robotics},
  volume = {8},
  number = {82},
  pages = {eadg1462},
  year = {2023},
  doi = {10.1126/scirobotics.adg1462}}

@ARTICLE{sunComparativeStudyNonlinear2022,
  author={Sun, Sihao and Romero, Angel and Foehn, Philipp and Kaufmann, Elia and Scaramuzza, Davide},
  journal={IEEE Transactions on Robotics}, 
  title={A Comparative Study of Nonlinear MPC and Differential-Flatness-Based Control for Quadrotor Agile Flight}, 
  year={2022},
  volume={38},
  number={6},
  pages={3357-3373},
  doi={10.1109/TRO.2022.3177279}}

@ARTICLE{sunFastUAVTrajectory2021,
  title = {Fast {UAV} Trajectory Optimization Using Bilevel Optimization with Analytical Gradients},
  author = {Sun, Weidong and Tang, Gao and Hauser, Kris},
  year={2021},
  journal={arXiv preprint arXiv:1811.10753},
  eprintclass = {cs},
  urldate = {2023-07-18},
  pubstate = {preprint}
}

@ARTICLE{zhouEfficientRobustTimeOptimal2023,
  author={Zhou, Ziyu and Wang, Gang and Sun, Jian and Wang, Jikai and Chen, Jie},
  journal={IEEE Robotics and Automation Letters}, 
  title={Efficient and Robust Time-Optimal Trajectory Planning and Control for Agile Quadrotor Flight}, 
  year={2023},
  volume={8},
  number={12},
  pages={7913-7920},
  doi={10.1109/LRA.2023.3322075}}

@ARTICLE{loquercioAutoTuneControllerTuning2022,
  author={Loquercio, Antonio and Saviolo, Alessandro and Scaramuzza, Davide},
  journal={IEEE Robotics and Automation Letters}, 
  title={AutoTune: Controller Tuning for High-Speed Flight}, 
  year={2022},
  volume={7},
  number={2},
  pages={4432-4439},
  doi={10.1109/LRA.2022.3146897}}

@ARTICLE{JiaDragUtil2022,
  author={Jia, Jindou and Guo, Kexin and Yu, Xiang and Zhao, Weihua and Guo, Lei},
  journal={IEEE Robotics and Automation Letters}, 
  title={Accurate High-Maneuvering Trajectory Tracking for Quadrotors: A Drag Utilization Method}, 
  year={2022},
  volume={7},
  number={3},
  pages={6966-6973},
  doi={10.1109/LRA.2022.3176449}}

@INPROCEEDINGS{mellingerMinimumSnapTrajectory2011,
  author={Mellinger, Daniel and Kumar, Vijay},
  booktitle={2011 IEEE International Conference on Robotics and Automation ({ICRA})}, 
  title={Minimum snap trajectory generation and control for quadrotors}, 
  year={2011},
  volume={},
  number={},
  pages={2520-2525},
  doi={10.1109/ICRA.2011.5980409}}

@ARTICLE{faesslerDifferentialFlatnessQuadrotor2018,
  author={Faessler, Matthias and Franchi, Antonio and Scaramuzza, Davide},
  journal={IEEE Robotics and Automation Letters}, 
  title={Differential Flatness of Quadrotor Dynamics Subject to Rotor Drag for Accurate Tracking of High-Speed Trajectories}, 
  year={2018},
  volume={3},
  number={2},
  pages={620-626},
  doi={10.1109/LRA.2017.2776353}}

@INPROCEEDINGS{morrellDifferentialFlatnessTransformations2018a,
  title = {Differential Flatness Transformations for Aggressive Quadrotor Flight},
  booktitle = {2018 {{IEEE International Conference}} on {{Robotics}} and {{Automation}} ({{ICRA}})},
  author = {Morrell, Benjamin and Rigter, Marc and Merewether, Gene and Reid, Robert and Thakker, Rohan and Tzanetos, Theodore and Rajur, Vinay and Chamitoff, Gregory},
  year={2018},
  pages = {5204--5210},
  issn = {2577-087X},
  doi = {10.1109/ICRA.2018.8460838},
  urldate = {2024-01-27},
  eventtitle = {2018 {{IEEE International Conference}} on {{Robotics}} and {{Automation}} ({{ICRA}})}
}

@Article{Andersson2019,
  author = {Joel A E Andersson and Joris Gillis and Greg Horn
            and James B Rawlings and Moritz Diehl},
  title = {{CasADi} - {A} software framework for nonlinear optimization
           and optimal control},
  journal = {Mathematical Programming Computation},
  volume = {11},
  number = {1},
  pages = {1--36},
  year = {2019},
  publisher = {Springer},
  doi = {10.1007/s12532-018-0139-4}
}

@ARTICLE{torrenteDataDrivenMPCQuadrotors2021,
  title = {Data-Driven {{MPC}} for Quadrotors},
  author = {Torrente, Guillem and Kaufmann, Elia and F{\"o}hn, Philipp and Scaramuzza, Davide},
  year = {2021},
  journal = {IEEE Robotics and Automation Letters},
  volume = {6},
  number = {2},
  pages = {3769--3776},
  issn = {2377-3766},
  doi = {10.1109/LRA.2021.3061307}
}

@ARTICLE{jiaEVOLVEROnlineLearning2024,
  title = {{{EVOLVER}}: Online Learning and Prediction of Disturbances for Robot Control},
  shorttitle = {{{EVOLVER}}},
  author = {Jia, Jindou and Zhang, Wenyu and Guo, Kexin and Wang, Jianliang and Yu, Xiang and Shi, Yang and Guo, Lei},
  year = {2024},
  journal = {IEEE Transactions on Robotics},
  volume = {40},
  pages = {382--402},
  issn = {1941-0468},
  doi = {10.1109/TRO.2023.3326318},
  urldate = {2024-02-14},
  eventtitle = {{{IEEE Transactions}} on {{Robotics}}}
}

@INPROCEEDINGS{bansalLearningQuadrotorDynamics2016,
  author={Bansal, Somil and Akametalu, Anayo K. and Jiang, Frank J. and Laine, Forrest and Tomlin, Claire J.},
  booktitle={2016 IEEE 55th Conference on Decision and Control (CDC)}, 
  title={Learning quadrotor dynamics using neural network for flight control}, 
  volume={},
  number={},
  pages={4653-4660},
  year={2016},
  doi={10.1109/CDC.2016.7798978}}

@ARTICLE{salzmannRealtimeNeuralMPCDeep2023,
  author={Salzmann, Tim and Kaufmann, Elia and Arrizabalaga, Jon and Pavone, Marco and Scaramuzza, Davide and Ryll, Markus},
  journal={IEEE Robotics and Automation Letters}, 
  title={Real-Time {neural MPC}: Deep Learning Model Predictive Control for Quadrotors and Agile Robotic Platforms}, 
  year={2023},
  volume={8},
  number={4},
  pages={2397-2404},
  doi={10.1109/LRA.2023.3246839}}

@ARTICLE{taoDiffTuneMPCClosedLoopLearning2023,
  author={Tao, Ran and Cheng, Sheng and Wang, Xiaofeng and Wang, Shenlong and Hovakimyan, Naira},
  journal={IEEE Robotics and Automation Letters}, 
  title={{{DiffTune-MPC}}: Closed-Loop Learning for Model Predictive Control}, 
  year={2024},
  volume={9},
  number={8},
  pages={7294-7301},
  doi={10.1109/LRA.2024.3422836}}

@INPROCEEDINGS{chenNeuralOrdinaryDifferential2019,
author = {Chen, Ricky T. Q. and Rubanova, Yulia and Bettencourt, Jesse and Duvenaud, David},
title = {Neural ordinary differential equations},
year = {2018},
publisher = {Curran Associates Inc.},
booktitle = {32nd International Conference on Neural Information Processing Systems (NeuraIPS)},
pages = {6572–6583},
}

@ARTICLE{richardsAdaptiveControlOrientedMetaLearningNonlinear2021,
	title={Control-oriented meta-learning},
	author={Richards, Spencer M and Azizan, Navid and Slotine, Jean-Jacques and Pavone, Marco},
	journal={The International Journal of Robotics Research},
	volume={42},
	number={10},
	pages={777--797},
	year={2023}
}

@ARTICLE{brunton2021modernkoopmantheorydynamical,
      title={Modern Koopman Theory for Dynamical Systems}, 
      author={Steven L. Brunton and Marko Budišić and Eurika Kaiser and J. Nathan Kutz},
      journal={arXiv preprint arXiv:2102.12086},
      year={2021},
      eprint={2102.12086},
      archivePrefix={arXiv},
      primaryClass={math.DS},
}

@INPROCEEDINGS{jackson2023dataefficientmodellearningcontrol,
  title = 	 {Data-Efficient Model Learning for Control with Jacobian-Regularized Dynamic-Mode Decomposition},
  author =       {Jackson, Brian Edward and Lee, Jeong Hun and Tracy, Kevin and Manchester, Zachary},
  booktitle = 	 {6th Conference on Robot Learning (CoRL)},
  pages = 	 {2273--2283},
  volume = 	 {205},
  year = 	 {2023}}

@INPROCEEDINGS{bauersfeldNeuroBEMHybridAerodynamic2021,
  title = {{{NeuroBEM}}: Hybrid Aerodynamic Quadrotor Model},
  shorttitle = {{{NeuroBEM}}},
  booktitle = {Robotics: {{Science}} and {{Systems XVII}} (RSS)},
  author = {Bauersfeld, Leonard and Kaufmann, Elia and Foehn, Philipp and Sun, Sihao and Scaramuzza, Davide},
  year = {2021},
  eprint = {2106.08015},
  primaryclass = {cs},
  doi = {10.15607/RSS.2021.XVII.042},
  urldate = {2024-07-16},
  archiveprefix = {arXiv}
}

@ARTICLE{savioloPhysicsInspiredTemporalLearning2022a,
  title = {Physics-Inspired Temporal Learning of Quadrotor Dynamics for Accurate Model Predictive Trajectory Tracking},
  author = {Saviolo, Alessandro and Li, Guanrui and Loianno, Giuseppe},
  year = {2022},
  journal = {IEEE Robotics and Automation Letters},
  volume = {7},
  number = {4},
  pages = {10256--10263},
  issn = {2377-3766},
  doi = {10.1109/LRA.2022.3192609},
  urldate = {2024-10-21}
}

@ARTICLE{kingmaAdamMethodStochastic2017,
      title={Adam: A Method for Stochastic Optimization}, 
      author={Diederik P. Kingma and Jimmy Ba},
      journal={arXiv preprint arXiv:1412.6980},
      year={2014},
      eprint={1412.6980},
      archivePrefix={arXiv},
      primaryClass={cs.LG},
}

@INPROCEEDINGS{yangtrace2024,
  author={Yang, Zihan and Jia, Jindou and Liu, Yuhang and Guo, Kexin and Yu, Xiang and Guo, Lei},
  booktitle={2024 IEEE 18th International Conference on Control and Automation (ICCA)}, 
  title={{TRACE}: Trajectory Refinement with Control Error Enables Safe and Accurate Maneuvers}, 
  year={2024},
  volume={},
  number={},
  pages={154-161},
  doi={10.1109/ICCA62789.2024.10591858}}

@ARTICLE{tase_nn_NARX,
  author={Xie, Jing and Bonassi, Fabio and Scattolini, Riccardo},
  journal={IEEE Transactions on Automation Science and Engineering}, 
  title={Learning Control Affine Neural {NARX} Models for Internal Model Control Design}, 
  year={2024},
  volume={},
  number={},
  pages={1-13},
  doi={10.1109/TASE.2024.3479321}}

@ARTICLE{tase_nn_MPC,
  author={Menegatti, Danilo and Giuseppi, Alessandro and Pietrabissa, Antonio},
  journal={IEEE Transactions on Automation Science and Engineering}, 
  title={Tractable Data-Driven Model Predictive Control Using One-Step Neural Networks Predictors}, 
  year={2024},
  volume={},
  number={},
  pages={1-12},
  doi={10.1109/TASE.2024.3453668}}

@ARTICLE{tase_learnCtrl,
  author={Wei, Mingxin and Zheng, Lanxiang and Wu, Ying and Liu, Han and Cheng, Hui},
  journal={IEEE Transactions on Automation Science and Engineering}, 
  title={Safe Learning-Based Control for Multiple UAVs Under Uncertain Disturbances}, 
  year={2024},
  volume={21},
  number={4},
  pages={7349-7362},
  doi={10.1109/TASE.2023.3341470}}

@INPROCEEDINGS{sun2024SafeStabilizationModel,
  author={Li, Ming and Sun, Zhiyong},
  booktitle={2024 IEEE 18th International Conference on Control and Automation (ICCA)}, 
  title={Safe Stabilization with Model Uncertainties: A Universal Formula with Gaussian Process Learning}, 
  year={2024},
  volume={},
  number={},
  pages={180-185}}

@INPROCEEDINGS{longhorizon2024,
  author={Rao, Pratyaksh Prabhav and Saviolo, Alessandro and Ferrari, Tommaso Castiglione and Loianno, Giuseppe},
  booktitle={2024 IEEE/RSJ International Conference on Intelligent Robots and Systems (IROS)}, 
  title={Learning Long-Horizon Predictions for Quadrotor Dynamics}, 
  year={2024},
  volume={},
  number={},
  pages={12758-12765},
  doi={10.1109/IROS58592.2024.10801793}}

@article{oconnell2022neuralfly,
author = {Michael O'Connell  and Guanya Shi  and Xichen Shi  and Kamyar Azizzadenesheli  and Anima Anandkumar  and Yisong Yue  and Soon-Jo Chung },
title = {Neural-Fly enables rapid learning for agile flight in strong winds},
journal = {Science Robotics},
volume = {7},
number = {66},
pages = {eabm6597},
year = {2022}}

@inproceedings{huangDATTDeepAdaptive2023,
  title = {{{DATT}}: Deep Adaptive Trajectory Tracking for Quadrotor Control},
  shorttitle = {{{DATT}}},
  booktitle = {7th {{Annual Conference}} on {{Robot Learning}}},
  author = {Huang, Kevin and Rana, Rwik and Spitzer, Alexander and Shi, Guanya and Boots, Byron},
  year = {2023},
}

@book{l1_adapt_ctrl,
author = {Hovakimyan, Naira and Cao, Chengyu},
title = {L1 Adaptive Control Theory},
publisher = {Society for Industrial and Applied Mathematics},
year = {2010},
}

@article{guo_antidisturbance_2014,
  author = {Lei Guo and Songyin Cao},
  title = {Anti-disturbance control theory for systems with multiple disturbances: A survey},
  journal = {ISA Transactions},
  volume = {53},
  number = {4},
  pages = {846-849},
  year = {2014},
  note = {Disturbance Estimation and Mitigation},
}

@ARTICLE{KNODE_MPC,
  author={Chee, Kong Yao and Jiahao, Tom Z. and Hsieh, M. Ani},
  journal={IEEE Robotics and Automation Letters}, 
  title={{KNODE-MPC}: A Knowledge-Based Data-Driven Predictive Control Framework for Aerial Robots}, 
  year={2022},
  volume={7},
  number={2},
  pages={2819-2826},
  doi={10.1109/LRA.2022.3144787}}

@ARTICLE{MillimeterLevel_PegInHole_2024,
	author={Wang, Meng and Chen, Zeshuai and Guo, Kexin and Yu, Xiang and Zhang, Youmin and Guo, Lei and Wang, Wei},
	journal={IEEE Transactions on Robotics}, 
	title={Millimeter-Level Pick and Peg-in-Hole Task Achieved by Aerial Manipulator}, 
	year={2024},
	volume={40},
	number={},
	pages={1242-1260}}

@ARTICLE{CompositeDisturbance_AerialManip_2025,
	author={Wang, Meng and Lyu, Shangke and Liu, Qianyuan and Yang, Ziqi and Guo, Kexin and Yu, Xiang},
	journal={IEEE Transactions on Automation Science and Engineering}, 
	title={Precise end-effector control for an aerial manipulator under composite disturbances: Theory and experiments}, 
	year={2025},
	volume={22},
	number={},
	pages={4006-4021}}

@article{JacobianUnknown_Lyu2020,
title = {Data-driven learning for robot control with unknown Jacobian},
author = {Shangke Lyu and Chien Chern Cheah},
journal = {Automatica},
volume = {120},
pages = {109120},
year = {2020},
issn = {0005-1098}}

@ARTICLE{nnMPC_AutonomousDriving2022,
  author={Spielberg, Nathan A. and Brown, Matthew and Gerdes, J. Christian},
  journal={IEEE Transactions on Control Systems Technology}, 
  title={Neural Network Model Predictive Motion Control Applied to Automated Driving With Unknown Friction}, 
  year={2022},
  volume={30},
  number={5},
  pages={1934-1945}}

\end{document}